\documentclass[11pt]{article}

\usepackage[margin=1in]{geometry}

\usepackage[authoryear,round]{natbib}
\setcitestyle{authoryear,open={(},close={)},citesep={; },aysep={,},yysep={, }}

\usepackage[usenames,dvipsnames]{color}
\usepackage{xcolor}
\definecolor{customred}{HTML}{E4312B}
\definecolor{customgreen}{HTML}{149954}

\usepackage{fixme}
\fxusetheme{color}
\fxsetup{status=draft} 

\usepackage{tabularx}
\usepackage{multirow}
\usepackage{booktabs}
\usepackage[normalem]{ulem}

\usepackage{listings}
\lstdefinelanguage{json}{
    basicstyle=\small\ttfamily,
    showstringspaces=false,
    breaklines=true,
    morecomment=[l]{//},
    morestring=[b]"
}

\usepackage{tcolorbox}
\tcbuselibrary{listings}

\usepackage[utf8]{inputenc}
\usepackage{xurl}
\usepackage{graphicx} 
\usepackage{hyperref} 

\title{Knowledge Graphs Generation from Cultural Heritage Texts: Combining LLMs and Ontological Engineering for Scholarly Debates}

\author{
Andrea Schimmenti\textsuperscript{1}, Valentina Pasqual\textsuperscript{1}, Fabio Vitali\textsuperscript{1}, Marieke van Erp\textsuperscript{2}\\
\\
\textsuperscript{1}Universit\`{a} degli Studi di Bologna, Bologna, Italy\\
\texttt{\{andrea.schimmenti2, valentina.pasqual2, fabio.vitali\}@unibo.it}\\
\\
\textsuperscript{2}KNAW Humanities Cluster, Amsterdam, the Netherlands\\
\texttt{marieke.van.erp@dh.huc.knaw.nl}
}

\date{} 

\begin{document}

\maketitle

\begin{abstract}
Cultural Heritage texts contain rich knowledge that is difficult to query systematically due to the challenges of converting unstructured discourse into structured Knowledge Graphs (KGs). This paper introduces ATR4CH (Adaptive Text-to-RDF for Cultural Heritage), a systematic five-step methodology for Large Language Model-based Knowledge Extraction from Cultural Heritage documents. We validate the methodology through a case study on authenticity assessment debates.\\
\textbf{Methodology} - ATR4CH combines annotation models, ontological frameworks, and LLM-based extraction through iterative development: foundational analysis, annotation schema development, pipeline architecture, integration refinement, and comprehensive evaluation. We demonstrate the approach using Wikipedia articles about disputed items (documents, artifacts...), implementing a sequential pipeline with three LLMs (Claude Sonnet 3.7, Llama 3.3 70B, GPT-4o-mini).\\
\textbf{Findings} - The methodology successfully extracts complex Cultural Heritage knowledge: 0.96-0.99 F$_1$ for metadata extraction, 0.7-0.8 F$_1$ for entity recognition, 0.65-0.75 F$_1$ for hypothesis extraction, 0.95-0.97 for evidence extraction, and 0.62 G-EVAL for discourse representation. Smaller models performed competitively, enabling cost-effective deployment.\\
\textbf{Originality} - This is the first systematic methodology for coordinating LLM-based extraction with Cultural Heritage ontologies. ATR4CH provides a replicable framework adaptable across CH domains and institutional resources.\\
\textbf{Research Limitations} - The produced KG is limited to Wikipedia articles. While the results are encouraging, human oversight is necessary during post-processing.\\
\textbf{Practical Implications} - ATR4CH enables Cultural Heritage institutions to systematically convert textual knowledge into queryable KGs, supporting automated metadata enrichment and knowledge discovery.
\end{abstract}


\section{Introduction}

Knowledge Graphs (KGs) have become the standard approach for representing and sharing Cultural Heritage (CH) information in the Linked Open Data (LOD) ecosystem, enabling interoperability between Libraries, Archives, and Museums institutions \citep{barabucci2021}. This effort has been mostly concentrated on creating KGs of metadata, with diversified workflows dedicated to converting semi-structured or already structured sources (catalogues, inventories) into LOD \citep{Bernasconi_Ferilli_2024}. However, the rich knowledge contained in unstructured CH texts — including descriptive content, contextual information, and analytical discourse — remains difficult to systematically extract and structure into queryable formats, and even when integrated into KGs, it is usually kept into long and description string fields \citep{barabucci2021,expliciting2025Giagnolini}. Scholarly authenticity assessment debates exemplify this challenge, where complex interpretative knowledge is embedded in natural language discourse but practically absent from structured representations. Alongside the overarching challenges, additional ones stem from the inherently interpretative nature of humanities scholarship, which aligns with a constructivist epistemology viewing knowledge as situated, provisional, and shaped by the observer's perspective. Checkland and Holwell \citep{checkland_data_2006} distinguish between \textit{data}—passively recorded facts—and \textit{capta}—knowledge actively constructed by the observer. This distinction challenges the realist assumption that often underpins data practices, in which data is treated as an objective and context-independent representation of reality.

These epistemological tensions manifest across various forms of CH scholarship, from attribution studies and provenance research to interpretative analysis and critical evaluation. Historical authenticity assessment exemplifies these challenges, as scholars from different humanities disciplines (e.g. Diplomatics, Palaeography, Philology, History) and scientific fields (e.g. Forensics, Materials science, Chemical analysis) frequently arrive at divergent conclusions based on different evidential priorities \citep{barone1912}. Inherent factors contributing to this diversity include historical uncertainty, gaps in documentary transmission, and subjectivity \citep{blau2011,gadamer2013}. 

Recent theoretical advancements acknowledge the subjectivity and uncertainty inherent in interpreting CH data, recognizing these as essential epistemic characteristics that must be preserved in digital representations \citep{amsdottoratoPasqual,piotrowski2020prospects,piotrowski2023uncertainty}.

Despite these conceptual advances, current KG implementations represent only simplified versions of scholarly discourse. This representational gap between rich textual discourse and sparse structured data is systematic across CH domains. Whether dealing with artistic attribution, provenance disputes, historical interpretation, or authenticity assessment, complex scholarly reasoning gets reduced to simple categorical assertions. Major knowledge bases like Wikidata\footnote{\url{https://www.wikidata.org/}} and DBpedia\footnote{\url{https://www.dbpedia.org/}} exemplify this limitation. While Wikipedia articles contain rich discussions with detailed scholarly arguments, evidence analysis, and alternative hypotheses, their structured counterparts reduce this complexity to sparse, categorical statements that fail to capture the evidential reasoning, methodological disagreements, and evolving consensus that characterize authentic scholarly discourse.

This pattern is systematic across CH materials: complex scholarly debates about document authenticity, artistic attribution, or historical interpretation are reduced to simple categorical assertions like "historical forgery" or "attributed to X," stripping away the evidential reasoning, competing hypotheses, and methodological considerations that form the substance of scholarly discourse. While Wikidata employs a custom reification method to integrate claims with varying degrees of truthfulness through its ranking mechanism, annotators in the CH domain sometimes neglet this feature \citep{dipasquale2024}. Consider the so-called \textit{Donation of Constantine}, a supposed 4th-century decree by Emperor Constantine transferring authority over Rome and the western Roman Empire to the Pope. In the 15th century, Lorenzo Valla exposed the document as a forgery through philological analysis \citep{valla2023}, demonstrating that its Latin contained anachronisms from the 8th rather than 4th century. Despite Valla's compelling evidence, acceptance of this finding evolved gradually over centuries.

\begin{figure}[th]
\centering
\includegraphics[width=0.8\linewidth]{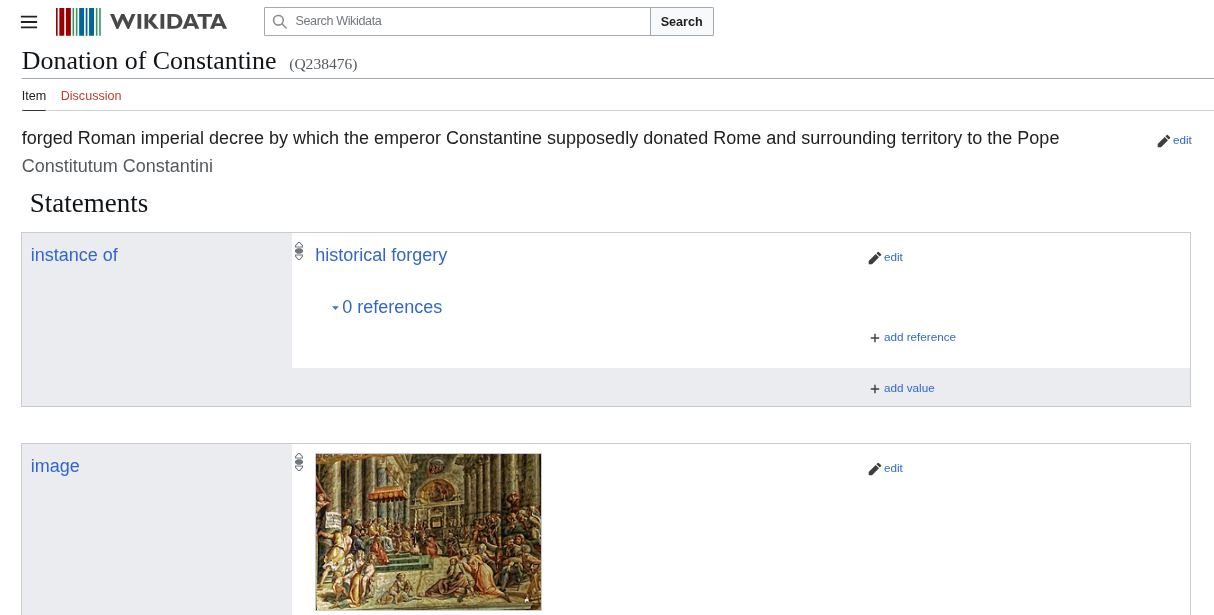}
\caption{The Donation of Constantine entry in Wikidata}
\label{fig:example_donation_wikidata}
\end{figure}

As shown in Figure~\ref{fig:example_donation_wikidata}, Wikidata categorizes the \textit{Donation} as a "historical forgery"\footnote{ \href{https://www.wikidata.org/entity/Q238476}{Donation of Constantine - Q238476}} with no representation of the scholarly debate, while DBpedia\footnote{\href{https://dbpedia.org/page/Donation_of_Constantine}{Donation of Constantine - DBpedia entry}} similarly lacks structured representation of the authenticity discourse. In contrast, the corresponding Wikipedia page contains extensive discussions of Valla's philological arguments, the specific linguistic evidence, the Church's resistance, and subsequent scholarly confirmation. This fundamental \textbf{misalignment} between rich textual content and sparse structured claims illustrates the core challenge this research addresses.

This misalignment stems from two interconnected challenges. The first is \textit{syntactic}: representing competing scholarly opinions within formal knowledge representation systems requires sophisticated mechanisms that traditional implementations struggle to handle effectively \citep{amsdottoratoPasqual}. While theoretical frameworks such as RDF-star, Named Graphs, and reification methods provide the necessary expressive power, their practical application demands complex modeling decisions about contradictory evidence, evolving consensus, and methodological disagreements, often resulting in oversimplified categorical assertions or unmanageably complex representations.

The second challenge is \textit{practical}: extracting complex scholarly information from textual sources requires enormous manual labor, creating insurmountable scalability barriers. This process requires that expert annotators to identify scholarly agents, extract evidential reasoning, and capture alternative hypotheses while maintaining consistency across large document collections, a prohibitively expensive undertaking for most CH institutions that combines technical knowledge representation skills with deep humanities scholarship expertise.

CH institutions possess vast textual resources containing sophisticated scholarly analyses, but lack practical means to transform this knowledge into queryable, machine-readable formats. Large Language Models (LLMs) present a promising solution due to their sophisticated ability to parse complex academic discourse, identify implicit relationships, and handle domain-specific vocabularies \citep{trends2024KGs}. LLMs do not require large corpora of annotated data and can exploit transfer learning on different domains than those they were explicitly trained on \citep{fewshotlearners}.

\textbf{Research Questions.}\label{research-questions} This work tackles the following primary research question:
\textit{How can a systematic methodology coordinate LLM-based Knowledge Extraction (KE) with ontological frameworks to effectively capture and structure the complex interpretative knowledge contained in CH texts?}

To systematically address this primary question, we investigate the following sub-questions, which we answer using our authenticity assessment case study:
\begin{enumerate}
\item \textbf{Methodological Framework:} What methodological approach can effectively coordinate LLM-based KE with existing ontological frameworks to capture complex scholarly interpretations in CH texts?\label{rq:methodological-framework}
\item \textbf{Extraction Performance:} How accurately can systematic LLM-based pipelines extract different components of scholarly discourse, including metadata, agents, evidential reasoning, and interpretative hypotheses?\label{rq:extraction-performance}

\item \textbf{Representation Fidelity:} Do automatically generated Knowledge Graphs adequately represent the complexity and nuance of scholarly interpretations when following structured methodological approaches?\label{rq:representation-fidelity}

\item \textbf{Model Comparison:} How do different LLMs perform within structured extraction pipelines for CH texts, and what are the implications for cost-effective deployment?\label{rq:model-comparison}

\item \textbf{Methodology Validation:} What insights does authenticity assessment validation provide about the methodology's broader applicability to other forms of CH interpretative scholarship?\label{rq:deployment-implications}
\end{enumerate}

To address these research questions, we developed ATR4CH (Adaptive Text-to-RDF for Cultural Heritage), a systematic five-step methodology that combines annotation development, ontological alignment, and LLM-based extraction. We validate this methodology through authenticity assessment debates, a challenging domain that exemplifies the complex evidential reasoning, scholarly disagreement, and multi-perspectival structures characteristic of CH interpretative scholarship.

Our contributions are threefold. First, we present the ATR4CH methodology itself, providing a replicable framework for systematic LLM-based KE that can be adapted across CH domains and institutional resources. Second, we demonstrate the practical implementation by connecting ATR4CH to the SEBI ontology to develop annotation models and extraction pipelines for complex scholarly discourse. Third, we provide a comprehensive evaluation on a manually annotated sample of Wikipedia articles, establishing performance benchmarks across multiple extraction tasks and model architectures.

Our evaluation demonstrates the methodology's effectiveness, achieving F$_1$-scores of 0.96-0.99 for metadata extraction, 0.7-0.8 for scholarly entity recognition, 0.65-0.75 for hypothesis extraction, and 0.95-0.97 for evidence extraction, with 0.62 G-EVAL overall discourse representativeness. Notably, smaller models performed competitively with larger architectures, indicating cost-effective deployment potential for resource-constrained CH institutions.

The remainder of the paper is organized as follows: Section~\ref{sec:relwork} reviews related works in knowledge representation, extraction methods for the Semantic Web, opinion mining, and LLM-based KE approaches. Section~\ref{sec:methodology} presents the ATR4CH methodology, detailing our five-step iterative approach for coordinating LLM-based extraction with ontological frameworks for the CH domain. Section~\ref{sec:dataset-ontolology-annotation} describes our corpus, describes the SEBI ontology, and explains the development of our annotation model through INCEpTION. Section~\ref{sec:methodology_implementation} presents the implementation of LLM-based pipeline for our specific case. Section~\ref{sec:results} provides detailed experimental results across five evaluation questions, comparing the performance of Claude Sonnet 3.7, Llama 3.3 70B, and GPT-4o-mini on metadata extraction, entity recognition, evidence mining, hypothesis extraction, and overall KE fidelity. Finally, Section~\ref{sec:discussion} discusses our findings in relation to the presented research questions, analyzes performance trade-offs, addresses deployment implications, and outlines contributions, limitations, and future research directions. The methodology is designed to be adaptable across CH domains requiring extraction of multi-perspectival interpretative knowledge, with authenticity assessment serving as a validation case that demonstrates the approach's effectiveness on particularly complex scholarly discourse. To ensure reproducibility of this work, all the code is published on a GitHub repository (\url{https://anonymous.4open.science/r/SEBI-Knowledge-Extraction-5FCE}).


\section{Related Work}\label{sec:relwork}
The challenges of representing and extracting interpretative knowledge in the CH domain have received increasing attention in recent research, particularly concerning Knowledge Representation (KR) and extraction (KE) tasks. We focus first on conceptual and ontological models developed for multi-perspective KR, and then turn to methods for extracting such interpretations from unstructured texts, including recent advances in LLMs. KE is intimately connected to KR. One influences the other, and depending on the specific KR model adopted, KE practices must be adapted accordingly \citep{maynard2017natural}.

\subsection{Knowledge Representation}\label{subsec:knowledge_representation}
Recent theoretical advancements acknowledged the subjectivity and uncertainty inherent in interpreting CH data, recognizing these aspects as essential epistemic characteristics in analyzing and representing such data \citep{piotrowski2020prospects}. Uncertainty in the CH domain arises not only from the data itself (for example, data extracted from the digitisation of a birth certificate) but also from the interpretative connections made by scholars regarding such data, such as identifying a name on a birth certificate with a specific historical figure \citep{piotrowski2023uncertainty}. However, these advancements have not translated into widely adopted practical tools and standards in KGs. Linked Open Data (LOD) is the standard for encoding and publishing CH data on the Web, promoting interoperability and data exchange between institutions. Standard online catalogues (e.g., Europeana)\footnote{\url{https://www.europeana.eu/}} typically provide single-perspective flat metadata, relegating discussions, debates, and uncertain facts to free text descriptions \citep{barabucci2021}. 

To the best of our knowledge, Wikidata is the only large-scale data catalogue that employs a custom reification method to integrate claims with varying degrees of truthfulness, i.e., its ranking mechanism. Despite the adequate expressive power made available by the Wikidata model, annotators in the CH domain underutilise this feature. Additionally, claims related to CH data often make use of numerous qualifiers to encode contextual metadata, likely due to the increased effort required for this type of annotation \citep{dipasquale2024}.

Some ontologies have been designed to structure multi-perspective representations in CH data. ICON \citep{sartini2023, baronicini2023} encodes visual recognitions in art history using $n$-ary relations to encode contextual metadata. Digital Hermeneutics \citep{daquino2020} employs a layered approach using Named Graphs \citep{carroll2005} to represent scholarly interpretations in archival and literary sources. HiCo \citep{daquino2015} and the STAR model \citep{andrews2023} have been designed to represent historical interpretations and arguments. Wider adoption is hampered by the absence of tools to support the extraction, categorization, and contextualization of scholarly interpretations at scale, where practical effectiveness ultimately depends on the ability to extract such interpretations from unstructured sources. This is the gap the work presented in this paper aims to fill: it introduces a method and supporting infrastructure for the formalization and publication of interpretative claims within CH datasets, addressing the need for representations of scholarly discourse.

\subsection{Knowledge Extraction for the Semantic Web}
In the Semantic Web domain, KG generation can involve Graph Neural Networks (GNNs) capable of generating, enriching or correcting valid RDF from multiple sources. In the CH domain, KG generation is usually based on mappings from other sources, with some exceptions such as CIDOC2VEC \citep{info12120503} building similarity recommendations using recurrent paths in CIDOC-CRM based KGs. Generally, generating KGs from text is closely linked to NLP tasks of Named Entity Recognition (NER), Relationship Extraction (RE), Entity Linking (EL) and similar subtasks \citep{maynard2017natural}, commonly termed Text2KG.

Recent CH works demonstrate diverse Text2KG approaches. The Musical Meetups Knowledge Graph (MMKG) \citep{morales_tirado_musical_2023} leverages Wikipedia biographies and DBpedia Spotlight for entity recognition, using existing LOD as filters and enrichment sources. MusicBO \citep{gangemi_musicbo_2024} uses Text2AMR2FRED \citep{Meloni2017AMR2FREDAT}, employing Abstract Meaning Representation (AMR) mapping to RDF/OWL with BLEURT score quality control. The Odeuropa project \citep{lisena_capturing_2022} integrates annotation models with ontological representation through frame-based annotation schemes that map directly to CIDOC-CRM extended concepts, demonstrated through multilingual BERT-based models across seven European languages.

\subsection{Opinion Mining for the Semantic Web}
These approaches reveal three paradigms in CH Text2KG extraction: leveraging existing encyclopedic resources with focused ontologies (MMKG), employing sophisticated semantic parsing with quality validation (MusicBO), and designing integrated annotation-ontology frameworks (Odeuropa), underscoring the need for flexible methodologies addressing varying source types and resource constraints.

Aspect-Based Sentiment Analysis (ABSA) through SemEval 2014 \citep{pontiki2014semeval} established granular opinion extraction frameworks with four subtasks: aspect term extraction, aspect term polarity, aspect category detection, and aspect category polarity. Recent advancements expanded ABSA to diverse datasets \citep{dong2014adaptive,gao2019target,hamborg2021towards}, with context-aware language models such as BERT improving performance \citep{gao2019target} and LLMs such as GPT-3.5 achieving state-of-the-art results with zero-shot prompting \citep{wang2023ischatgpt}. An ABSA-oriented approach could be transferred to scholarly opinions within a debate, as  analyzing the sentiment and the specific aspects of a review is not structurally dissimilar to a scholarly opinion, where some evidences (parallel to aspects) are followed by an assertion (parallel to sentiment).

\subsection{LLMs for Knowledge Extraction}
LLMs have introduced new text-to-KG extraction paradigms \citep{trends2024KGs,mihindukulasooriya2023text2kgbench,meyer2024llm}. Allen et al. \citep{KE_using_LLMs} identify two primary directions: hybrid neuro-symbolic systems and natural language interfaces for domain experts—particularly relevant for CH contexts where experts lack technical expertise but possess deep interpretative knowledge.

Lairgi et al. \citep{lairgiItext2kg2024} address traditional pipeline limitations through iText2KG's zero-shot, incremental approach with four modules enabling dynamic knowledge base expansion. Kumar et al. \citep{kumar_k-lm_2022} demonstrate Knowledge Language Models (K-LM) injecting domain-specific RDF triples into language model architectures, revealing that relevance matters more than quantity of injected KGs. Ringwald \citep{ringwald_learning_2024} explores pattern-based extraction methods learning from Wikipedia-DBpedia/Wikidata pairs, offering middle ground between rigid rules and unpredictable LLM generation while maintaining alignment with established ontological frameworks.


\section{The ATR4CH Methodology}\label{sec:methodology}

This section introduces the \textit{Adaptive Text-to-RDF for Cultural Heritage} (ATR4CH) methodology, an iterative approach specifically designed for extracting KGs from documents using LLMs in the CH, and by extension, Humanities, domain. Unlike traditional methodologies that treat annotation, KE, and ontology alignment as separate processes, ATR4CH recognizes that these processes are fundamentally interdependent in humanities contexts and must therefore be approached not sequentially but conjunctively. 

\subsection{Methodology Overview}\label{subsec:methodology-overview}

\begin{figure}
    \centering
    \includegraphics[width=0.89\linewidth]{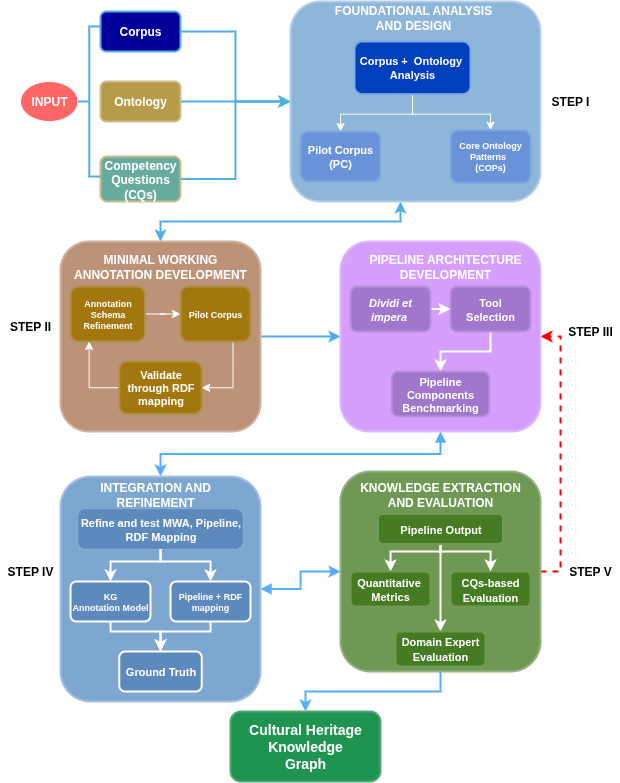}
    \caption{Flowchart of the ATR4CH methodology showing the five-step iterative process}
    \label{fig:methodology}
\end{figure}

The ATR4CH methodology systematically transforms three foundational inputs into a validated KE system. The process begins with a corpus of unstructured documents, a target ontology defining the desired knowledge representation, and a set of Competency Questions (CQs) specifying what the system should be able to answer. Through five sequential steps, these inputs are transformed into a working extraction pipeline, a refined annotation model, and a comprehensive evaluation framework.
A flowchart of the methodology is shown in Figure \ref{fig:methodology}.

The methodology is designed to converge on a solution that meets both technical KE requirements and ontological representation standards for CH information.

Several foundational approaches in ontology engineering and KE are behind this methodology. The eXtreme Design methodology \citep{extreme_design_ontology} provides the theoretical foundation for iterative and incremental development in ontology engineering, emphasizing the centrality of CQs throughout the development process, an approach successfully applied in CH contexts such as Viewsari, a KG of Giorgio Vasari's The Lives \citep{viewsariOndraszek}. Additional inspirations include \citep{tomasi2020} for the selection of relevant items during the development phase, and the integrated annotation-ontology methodology used in Odeuropa's annotation model for mentions of smell throughout CH documents \citep{lisena_capturing_2022}, which demonstrated how annotation schemas can be designed to map directly to ontological concepts within CH applications.

ATR4CH specifically focuses on coordinating an annotation and extraction pipeline with an ontology or combination of multiple ontologies in the CH domain, such as CIDOC-CRM \citep{Doerr_2003}, Dublin Core\footnote{\url{https://www.dublincore.org/}}, FRBR/FRBRoo \citep{ifla2017frbr}, HiCo \citep{daquino2015}, SKOS\footnote{\url{https://www.w3.org/TR/skos-reference/}}, and PROV-O \citep{lebo2013}. The methodology is suited on unstructured texts (informative, narrative, scholarly sources) and less for semi-structured documents such as catalogues or inventories, which are also common sources in the Libraries, Archives and Museum (LAM) domain. The ATR4CH methodology recognizes LLMs as particularly useful for their capabilities in performing KE through In-Context Learning (ICL) \citep{brown2020language}, Few-Shot, and Chain-of-Thought (CoT) strategies \citep{petroni2019language,lairgiItext2kg2024}, based on established practices of KE \citep{Tamaauskait2022DefiningAK}. 

ATR4CH assumes the existence of at least one document containing information that can be mapped to a given ontology, where this information must be annotated to answer a set of CQs. The approach is designed to be resource-adaptive, accommodating varying computational capabilities and project scales while maintaining consistent theoretical foundations.

\subsection{Foundational Analysis and Design (Step I)}\label{subsec:methodology-step1}

The first step establishes foundational understanding of the source material by analyzing both corpus and ontology to identify core patterns for KE. This analysis addresses potential data sparseness problems, which are common throughout Information Extraction tasks on unstructured texts. 

\begin{itemize}
    \item \textbf{Corpus Analysis}: This preliminary step examines how knowledge manifests throughout textual discourse, including linguistic patterns, discourse structures, and representational strategies. Key challenges include implicit mentions requiring contextual inference, long-distance dependencies where KG components are separated by substantial text spans, nested entities discussed through relational structures, and ambiguous references using nicknames or figures of speech. In the case of Wikipedia articles about forged CH items, this analysis reveals specific patterns in both structure and content. Structurally, it identifies which sections contain scholarly opinions versus out-of-scope debates, enabling focused extraction from high-density sections like "Scholarly analysis" while filtering out biographical or tangential material. Content-wise, it determines whether articles present complete scholarly reasoning or merely final judgments (e.g., "most scholars consider this a forgery" versus detailed evidential arguments), guiding the methodology toward sources with sufficient depth for comprehensive KE.

    \item \textbf{Ontology Analysis}: This analysis assesses which parts of the target ontology can be populated from source documents, examining alignment between the ontology's conceptual framework and available textual information. It identifies which ontological classes and properties have sufficient textual evidence for extraction, which relationships can be reliably inferred from corpus discourse patterns, and which elements may need omission due to lack of textual support. Competency Questions guide prioritization of ontological coverage based on research requirements. For instance, in the case of the Donation of Constantine, a relevant CQ would be "What are the latest scholars identifying the document as authentic?"

    \item \textbf{Core Ontological Patterns (COPs) Identification}: Based on corpus and ontology analyses, this process identifies essential Knowledge Graph patterns required to answer the CQs. Core Ontological Patterns represent the central ontological nodes and relationships that are both present in the corpus as extractable information and necessary for addressing the research questions. For example, in authenticity assessment debates, a typical pattern would be "Scholar X evaluates Feature Y of Document Z using Method M and concludes Authenticity Status S." The identification process involves: (1) assessing alignment between Competency Questions, ontological structures, and available textual content, (2) identifying patterns with sufficient textual evidence for reliable extraction, (3) prioritizing based on extractability feasibility and CQ relevance, and (4) selecting a manageable subset that forms the semantic backbone for KE.
    
    \item \textbf{Pilot Corpus Selection}: The Pilot Corpus is a set of representative documents from the whole corpus that serves as a development sandbox for annotation and KE exploration. The pilot corpus is not intended as a quantitatively representative sample but rather as a qualitative sample that must be linguistically, structurally, and epistemically representative while remaining manageable for intensive manual development work. Selection criteria focus on ensuring coverage of various linguistic patterns, discourse structures, and diverse manifestations of the target COPs. The size can be as small as 3-5 documents, depending on document length and complexity of information manifestation patterns detected during corpus analysis.
\end{itemize}

\subsection{Minimal Working Annotation Development (Step II)}\label{subsec:methodology-step2}

The second step consists of iterative cycles to develop the Minimal Working Annotation (MWA). The aim of this step is to model, enrich, refine, and validate the MWA over the pilot corpus, and test whether the annotated data can be successfully mapped to KGs that satisfy the COPs identified in Step~\ref{subsec:methodology-step1}.

\textbf{MWA Schema Development}: Starting from the COPs identified in Step~\ref{subsec:methodology-step1}, this process develops an annotation schema that captures the essential knowledge structures while remaining practical for both manual annotation and automated extraction. The schema design must account for the diverse ways knowledge manifests in the corpus, as identified during the corpus analysis, including both explicit textual mentions and information that requires inference or contextualization.

The MWA should prioritize simplicity and feasibility while ensuring adequate coverage of the COPs. The "minimal" aspect refers to including only those annotation elements that are necessary for extracting the identified ontological patterns, avoiding over-annotation that may complicate the extraction process without contributing to answering the CQs. If the COPs require complex semantic structures beyond simple triple patterns, the annotation schema should include appropriate mechanisms for representing these relationships in a way that can be reliably mapped to RDF (e.g., if the ontology relies on Named Graphs, reification, etc). 

\textbf{Knowledge Base Integration Strategy}: Knowledge base integration is an integral part of annotation processes because it enables consistent entity identification and vocabulary alignment between textual mentions and the target ontology. Since the COPs typically involve ontological individuals, entities, controlled vocabularies, or standardized terminologies defined within or referenced by the ontology, annotators need access to these resources to ensure that textual references are linked to the correct ontological entities. Without this integration, the same real-world entity might be annotated inconsistently across documents, preventing proper aggregation and reasoning in the final KG.

This integration, whether through building local vocabularies or leveraging external resources such as Wikidata or DBpedia, must be designed early in the annotation model development to establish clear protocols for entity linking and vocabulary alignment that will guide both manual annotation and automated extraction in Step~\ref{subsec:methodology-step3}. The choice between local and external knowledge bases depends on domain coverage, data quality requirements, and the specific entity types required by the COPs, with early integration ensuring that the annotation schema can consistently handle entity disambiguation, coreference resolution, and terminological standardization throughout the development process.

\textbf{Iterative Development Process}: The development follows a systematic cycle designed to ensure that the annotation schema can successfully produce RDF structures that satisfy the COPs:

\begin{enumerate}
    \item \textbf{Schema Design}: Develop initial annotation layers based on the COPs, incorporating knowledge base integration protocols through tagsets, controlled vocabularies, and standardized terminologies that align with the target ontology.    
    \item \textbf{Pilot Corpus Annotation}: Annotate the entire pilot corpus using the current iteration of the MWA to identify potential gaps, inconsistencies, or practical bottlenecks in the annotation process.
    \item \textbf{Mapping Validation}: Conduct preliminary mapping exercises from the annotated data to RDF format, testing whether the resulting KGs satisfy the COPs and adequately represent the semantic content of the source documents.
    \item \textbf{Schema Refinement}: Refine the annotation model based on issues identified during mapping validation, returning to previous steps as necessary to address fundamental problems with the annotation approach.
\end{enumerate}

These preliminary mapping exercises are crucial for validating that the annotation schema can produce the target knowledge structures. They serve as an early validation mechanism, ensuring that the annotation effort will ultimately generate RDF that satisfies the COPs before proceeding to automated extraction development.

\subsection{Pipeline Architecture Development (Step III)}\label{subsec:methodology-step3}

The third step designs and implements computational tools to automatically extract COPs from text using the MWA as target schema, addressing CH corpora's domain-specific characteristics and limited annotated training data.

\textbf{Task Decomposition and Architecture Design}: The KE task is designed around MWA elements, prioritizing based on COP semantic importance and accounting for information manifestation patterns from corpus analysis. This modular approach enables incremental extraction, where KG components are progressively identified through sequential processing, facilitating debugging and targeted optimization while minimizing error propagation through robust intermediate representations.

\textbf{Tool Selection Strategy}: Tool choice aligns with available resources and data characteristics:
\begin{itemize}
    \item \textbf{Low data, low resources}: API-based LLMs with few-shot prompting and rule-based entity linking
    \item \textbf{Moderate data, moderate resources}: Hybrid approaches combining pre-trained models with domain-specific fine-tuning
    \item \textbf{Large data, extensive resources}: Custom model training and ensemble methods
    \item \textbf{Large data, low resources}: Structured pipeline approaches leveraging smaller models with knowledge distillation
\end{itemize}

LLM-based approaches use structured output generation through JSON schemas \citep{schick2023toolformer,qin2024toollearningfoundationmodels} and ICL strategies \citep{brown2020language,min2022rethinking}, combined with specialized NER tools \citep{devlin2019bert} for precise span identification.

\textbf{Pipeline Implementation}: Development targets the MWA schema, integrating knowledge base resources and vocabulary standardization protocols through prompt integration or RAG \citep{rag2020Lewis}. Initial implementation focuses on basic functionality across all COPs before optimization.

\textbf{Benchmarking Approach}: Evaluation strategies range from basic (standard metrics on pilot corpus) to comprehensive (ablation studies and hybrid approach exploration) based on project constraints.

\textbf{Output}: An extraction pipeline capable of processing raw text and generating structured outputs following the MWA schema.

\subsection{Integration and Refinement (Step IV)}\label{subsec:methodology-step4}

The fourth step harmonizes the COPs (Step~\ref{subsec:methodology-step1}), MWA (Step~\ref{subsec:methodology-step2}), and pipeline (Step~\ref{subsec:methodology-step3}) into a coherent, end-to-end KE system, bringing the experimental pipeline into production-ready status.

\textbf{End-to-End Pipeline Testing}: Comprehensive testing over the pilot corpus processes documents from raw text to final KGs, revealing systematic issues including data sparseness patterns, inconsistent tool coverage across discourse types, and representation generation errors. The testing systematically evaluates performance across document types and semantic phenomena, with particular attention to error propagation through pipeline stages.

\textbf{MWA Refinement to KG-AM}: Based on testing results, the MWA evolves into a production-ready KG Annotation Model (KG-AM) suitable for both manual annotation and automated extraction. This may involve adding elements crucial for automated extraction such as coreference chains, disambiguation tags, or confidence indicators while maintaining backward compatibility with COPs.

\textbf{Mapping Algorithm Enhancement}: Preliminary mapping algorithms from Step~\ref{subsec:methodology-step2} are refined based on pipeline testing requirements, improving handling of complex semantic structures and adding validation using tools such as SHACL, OWL reasoners, SPARQLAnything \citep{sparqlAnything2024Asprino} and RML \citep{dimou_ldow_2014}. Error handling mechanisms manage extraction failures and partial results.

\subsection{Knowledge Extraction and Evaluation (Step V)}\label{subsec:methodology-step5}

The final validation phase employs technical validation and domain-expert evaluation to ensure knowledge structures accurately represent domain-specific discourse complexity, applying the refined system from Step~\ref{subsec:methodology-step4} to test data.

\textbf{Ground Truth Preparation}: Comprehensive Ground Truth (GT) creation using the KG-AM involves annotating a test dataset separate from the pilot corpus, covering all COPs from Step~\ref{subsec:methodology-step1} and applying mapping algorithms from Step~\ref{subsec:methodology-step4} to generate reference RDF.

\textbf{Knowledge Extraction}: Test datasets are processed through the complete pipeline under realistic deployment conditions, with systematic documentation of performance and failure modes.

\textbf{Multi-Level Evaluation}: Multiple complementary approaches address KG evaluation challenges:
\begin{itemize}
    \item \textbf{Technical Evaluation}: Component-level assessment using precision, recall, and F$_1$-score, plus coverage analysis for Competency Questions
    \item \textbf{Semantic Evaluation}: KG "rehydration" \citep{gardent2017creating,gangemi_musicbo_2024} enables comparison when structural alignment is impossible, using metrics like BLEU \citep{papineni2002bleu}, METEOR \citep{banerjee2005meteor}, BARTScore \citep{yuan2021bartscore}, CHRF+++ \citep{popovic2015chrf}, and G-EVAL \citep{geval-evaluation} as proposed by \citep{he-etal-2025-evaluating-improving}
    \item \textbf{Competency-Based Evaluation}: SPARQL query suites derived from original CQs, aligned with \citep{extreme_design_ontology} evaluation using tools like TestaLOD \citep{testalod2019Carriero}
\end{itemize}

\textbf{Domain Expert Validation}: Comprehensive review by domain specialists evaluates extraction quality and coherence, with rehydration technique enabling evaluation by experts without RDF expertise.

\textbf{Iteration Strategy}: Evaluation results may trigger returns to earlier steps: coverage issues to Step~\ref{subsec:methodology-step2} or Step~\ref{subsec:methodology-step4}, extraction bottlenecks to Step~\ref{subsec:methodology-step3}, or systematic errors requiring architectural restructuring in Step~\ref{subsec:methodology-step4}.


\section{Corpus, Ontology and Annotation Model}\label{sec:dataset-ontolology-annotation}
As stated in \ref{sec:methodology}, our methodology starts from a corpus, an ontology, and a set of CQs. This section discusses how the dataset was collected and which documents it contains (\ref{subsec:dataset}); the ontology (see \ref{subsec:ontology}); and the implementation of the first two steps of the methodology, namely Step I (see \ref{subsec:methodology-step1}), and Step II (see \ref{subsec:methodology-step2}).

\subsection{Dataset}\label{subsec:dataset}
Our dataset comprises Wikipedia articles focusing on historical forgeries, hoaxes, and authenticity controversies across CH. The collection was compiled using automated web scraping from Wikipedia's categorical organization system, targeting categories related to forgeries and authenticity debates.

We scraped articles from 15 distinct Wikipedia categories, retrieving full article text and inter-article link structures (sitelinks\footnote{\url{https://www.wikidata.org/wiki/Help:Sitelinks}}). The script selects both categorical pages\footnote{See for instance the Wikipedia Category "Forgery" \\(\url{https://en.wikipedia.org/wiki/Category:Forgery})} and standalone articles\footnote{See for instance the article describing the Donation of Constantine \\(\url{https://en.wikipedia.org/wiki/Donation\_of\_Constantine})}, storing each document with complete textual content and associated metadata including categorization and cross-references to related entities.

The initial selection covered 31 categories, spanning from Document\footnote{\url{https://en.wikipedia.org/wiki/Category:Document\_forgeries}} and Literary Forgeries\footnote{\url{https://en.wikipedia.org/wiki/Category:Literary\_forgeries}} to Historical Myths\footnote{\url{https://en.wikipedia.org/wiki/Category:Historical\_myths}}, Conspiracy Theories\footnote{\url{https://en.wikipedia.org/wiki/Category:Conspiracy\_theories}}, Pseudepigraphy (i.e. falsely attributed works, texts whose claimed author is not the true author, or works whose real author attributed it to a figure of the past)\footnote{\url{https://en.wikipedia.org/wiki/Category:Pseudepigraphy}} and Political forgery\footnote{\url{https://en.wikipedia.org/wiki/Category:Political\_forgery}}. Out of the total 1301 documents which  were retrieved, \footnote{The selection was performed on October 2024}, 16 categories and 717 articles were excluded, as they did not present any scholarly debate or they were not about a CH item (be it a document, an artifact, etc).  
The dataset encompasses 581 articles as shown in Table \ref{tab:dataset-categories}.
\begin{table}[h]
\centering
\caption{Distribution of articles across Wikipedia categories in the corpus}
\label{tab:dataset-categories}
\begin{tabular}{lc}
\toprule
\textbf{Category} & \textbf{Article Count} \\
\midrule
Literary forgeries & 138 \\
Pseudepigraphy & 65 \\
Old Testament pseudepigrapha & 60 \\
Forgery controversies & 58 \\
Archaeological forgeries & 52 \\
Musical hoaxes & 44 \\
Art forgers & 40 \\
Document forgeries & 33 \\
Ancient Greek pseudepigrapha & 28 \\
Political forgery & 26 \\
Religious hoaxes & 15 \\
Modern pseudepigrapha & 11 \\
Sculpture forgeries & 7 \\
Political forgeries & 2 \\
Shakespeare authorship question & 2 \\
\midrule
\textbf{Total} & \textbf{581} \\
\bottomrule
\end{tabular}
\end{table}

\begin{figure}
    \centering
    \includegraphics[width=0.8\linewidth]{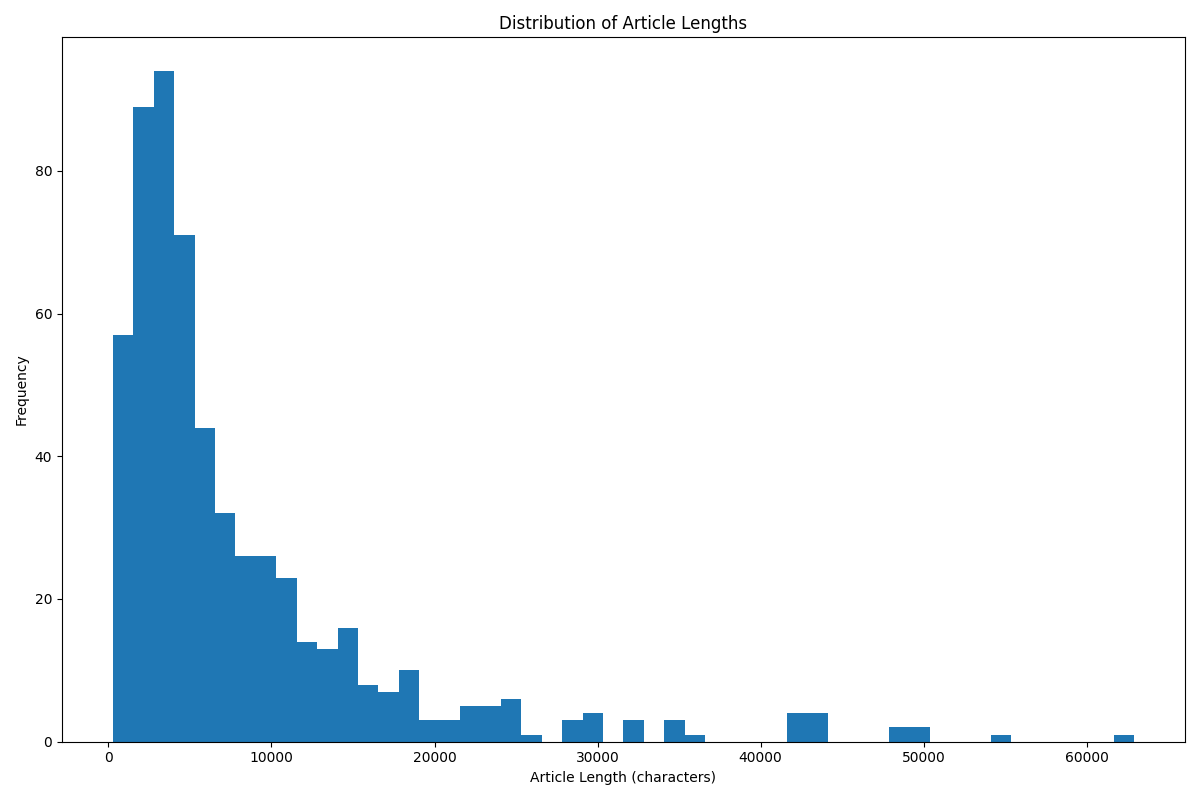}
    \caption{Overall distribution of article lengths showing the right-skewed pattern characteristic of encyclopedic content, with most articles in the 2k-15k character range and notable outliers extending beyond 40k characters.}
    \label{fig:article-length-distribution}
\end{figure}

The corpus exhibits significant variability in document length and complexity, with articles averaging 8,150 characters and 1,249 tokens per document. Unique vocabulary per article averages 464 tokens, indicating substantial lexical diversity within the scholarly discourse on authenticity assessment. As shown in Figure~\ref{fig:article-length-distribution}, the distribution of article lengths follows a right-skewed pattern, with most articles ranging from 2k to 15k characters, with some outliers that extend beyond 40k characters.

\begin{figure}
    \centering
    \includegraphics[width=0.8\linewidth]{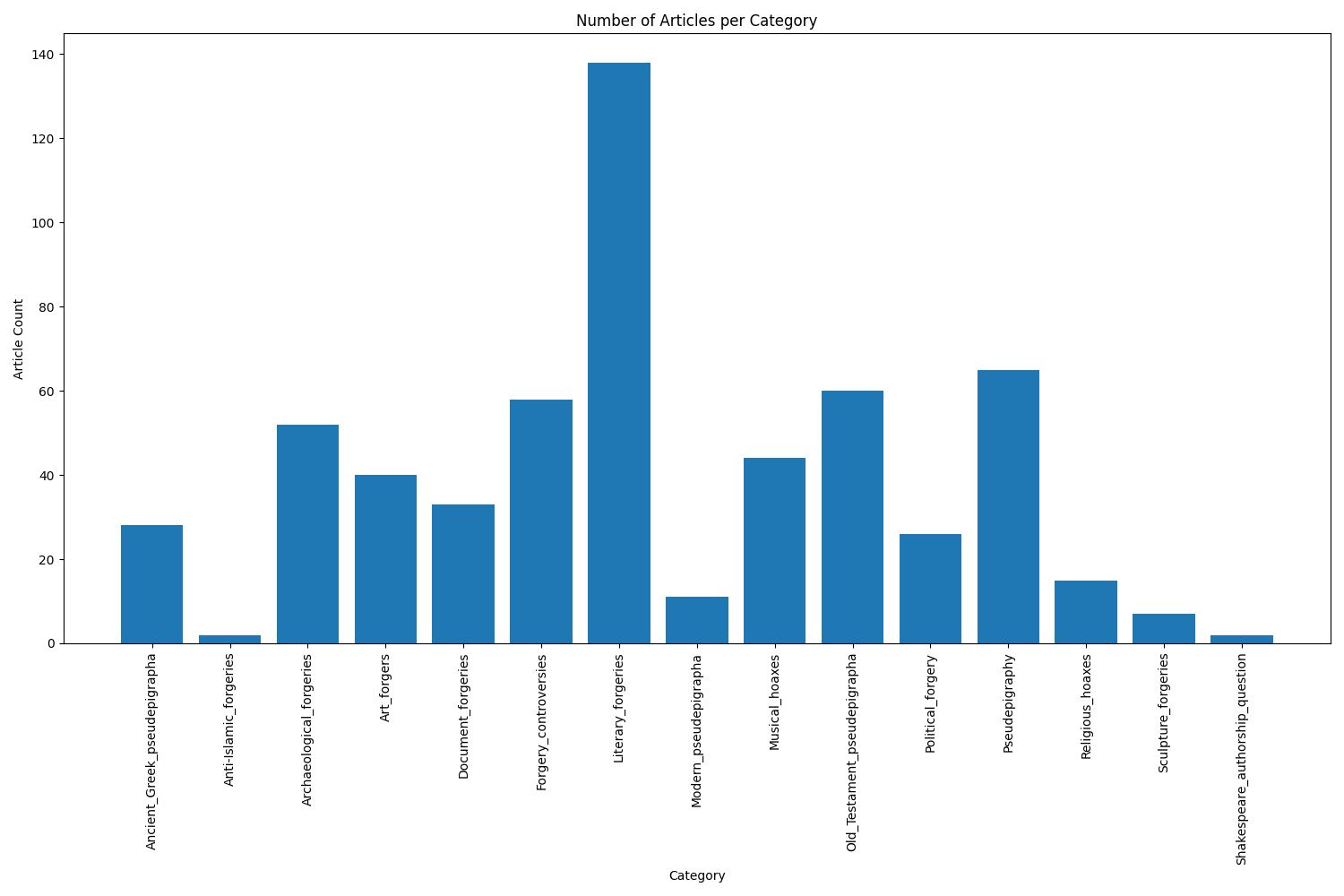}
    \caption{Distribution of articles across Wikipedia categories, showing the natural prevalence of different forgery types in scholarly discourse.}
    \label{fig:articles-per-category}
\end{figure}

The distribution across categories reflects the natural prevalence of different forgery types in scholarly discourse (Figure~\ref{fig:articles-per-category}). Literary forgeries represent the largest category with 138 articles, followed by various forms of pseudepigraphy totaling 136 articles across subcategories. Archaeological and artistic forgeries comprise 132 articles combined, while more specialized categories such as musical hoaxes and religious controversies contain fewer but often more detailed entries.

The temporal scope of the corpus spans from late antiquity to the contemporary period, representing diverse scholarly debates across different historical contexts.

\begin{figure}[t]
    \centering
    \includegraphics[width=0.8\linewidth]{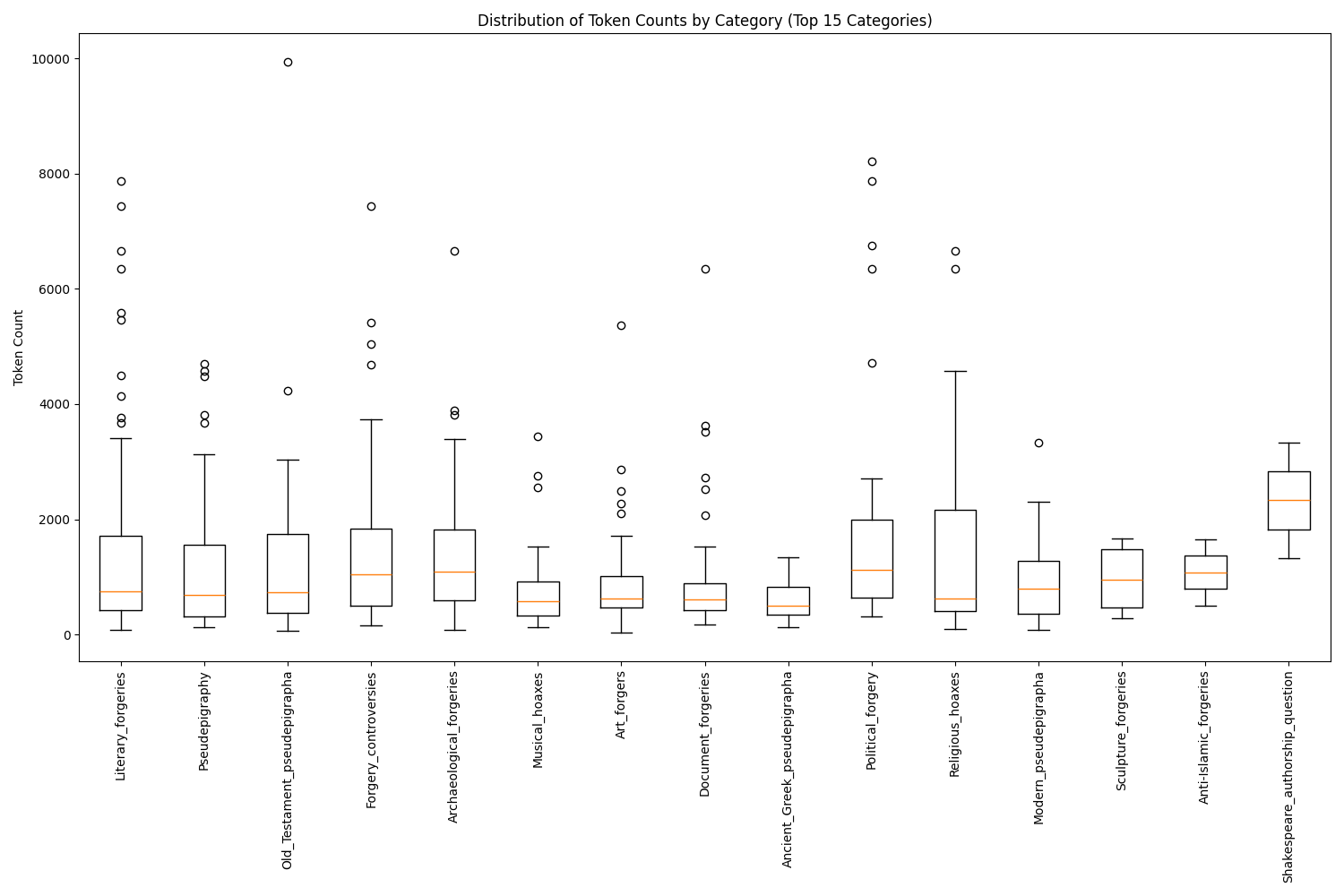}
    \caption{Token count distribution by category, illustrating variability in article length and content density. Box plots show medians, quartiles, and outliers representing comprehensive case studies.}
    \label{fig:token-distribution}
\end{figure}

Notable variations exist across categories in terms of content density and scope, as illustrated in Figure~\ref{fig:token-distribution}. The token distribution reveals substantial variability within categories, with numerous outliers indicating comprehensive case studies that warrant extensive coverage. The Shakespeare authorship question category demonstrates the highest token density, with articles reaching nearly 10K tokens, reflecting the extensive scholarly debate surrounding this topic. Political forgery and religious hoaxes also show elevated token counts, indicating rich semantic content suitable for KE tasks. Conversely, categories like musical hoaxes and modern pseudepigrapha exhibit more consistent, moderate-length articles with fewer outliers.

\begin{figure}[t]
\centering
\includegraphics[width=0.8\linewidth]{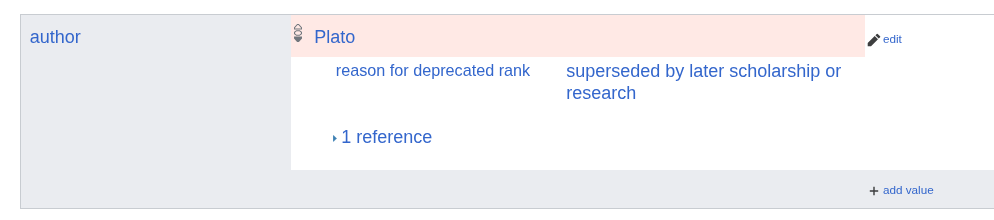}
\caption{Plato noted as the author of the Demodocus using a deprecated rank, illustrating how existing knowledge bases can represent disputed attributions}
\label{fig:demodocus_ewa}
\end{figure}

Among the corpus, one of the oldest CH items under discussion is the Demodocus\footnote{\url{https://en.wikipedia.org/wiki/Demodocus_(dialogue)}}, a fabricated Platonic dialogue that exemplifies early pseudepigraphic practices. For this specific case, a Wikidata entry exists\footnote{\url{https://www.wikidata.org/wiki/Q2625856}} which appropriately employs a deprecated rank for the authorship claim linking Plato to the Demodocus, as shown in Figure \ref{fig:demodocus_ewa}, as exemplified in Section \ref{subsec:knowledge_representation}.

Another article of the corpus is the Protocols of the Elders of Zion, one of the most famous forgeries to date\footnote{\url{https://en.wikipedia.org/wiki/The_Protocols_of_the_Elders_of_Zion}}. Among the most recent examples, the 1996 \textit{Posthumous Diary}\footnote{\url{https://en.wikipedia.org/wiki/Posthumous_Diary}}, an allegedly forged collection of poems by Italian poet Eugenio Montale, which caused a participated debate in the Italian philology community.

\subsection{Ontology}\label{subsec:ontology}
The Scholarly Evidence Based Interpretation ontology\footnote{\url{https://valentinapasqual.github.io/sebi/}}) (SEBI) \citep{amsdottoratoPasqual} was based on a selection of scholarly articles, e.g. \citep{hartel2017}, a catalogue describing 153 known forgeries from Styria \citep{haider2022}, and several discussions with an expert Diplomatist. The data model represents authenticity assessment claims using RDF-star  \citep{hartig2017foundations} as a reification method to represent all (possibly concurrent) claim contents as well as their contextual information \citep{daquino2020}. The content of each claim provides the following basic information about the document: authenticity classification, date and place of creation, author, and the intention behind its creation. Additionally, contextual information about the claim is recorded, specifically the evidences collected by the scholar to reach a certain conclusion (using HiCo),\footnote{\url{https://marilenadaquino.github.io/hico/}} evidence-based assessments as well as the author of the claim and relevant bibliographic entries (using PROV-o).\footnote{\url{https://www.w3.org/TR/prov-o/}} 
RDF-star \citep{hartig2017foundations} has been chosen as the reification method to express both the claim contents and context, allowing the representation of the entire evaluation process conducted by scholars.

\begin{figure}[h]
    \centering
    \includegraphics[width=0.8\linewidth]{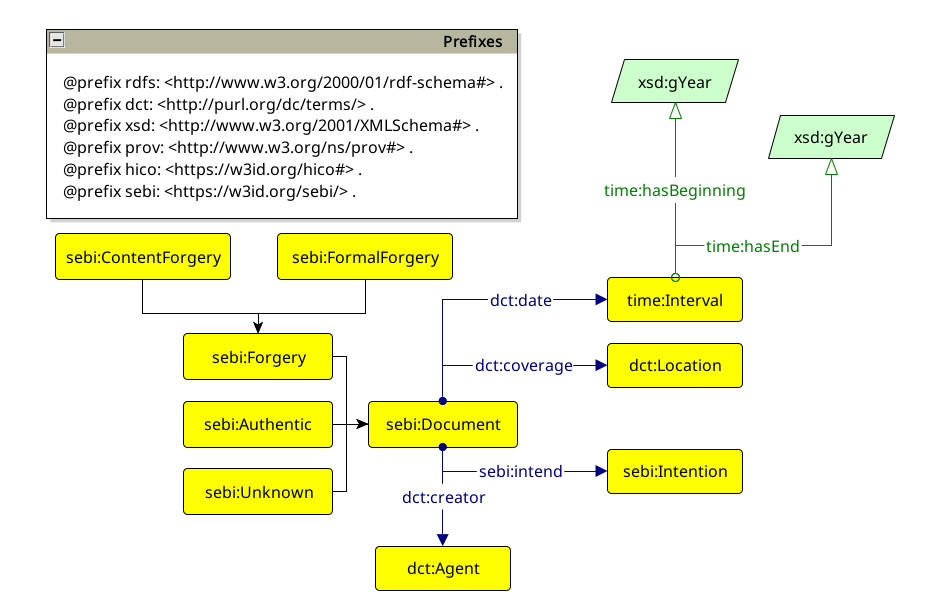}
    \caption{Selection of classes and properties to represent scholarly claims tackling authenticity assessment of a document}
    \label{fig:full_claim_datamodel}
\end{figure}

As shown in Figure \ref{fig:full_claim_datamodel} each claim contains an attempt of classification towards the CH item authenticity. This is obtained by making sure that all items are instances of one of the classes \texttt{sebi:Forgery}, \texttt{sebi:Authentic}, \texttt{sebi:FormalForgery}, \texttt{sebi:ContentForgery} all subclasses of \texttt{sebi:Document}.

Additionally, each RDF-star quoted triple includes details such as the believed creator of the document (expressed through \texttt{sebi:Document} - \texttt{dct:creator} - \texttt{dct:Agent}), the date of creation (\texttt{sebi:Document} - \texttt{dct:date} - \texttt{time:Interval}), location of creation (\texttt{sebi:Document} - \texttt{dct:coverage} - \texttt{dct:Location}) and the intention behind the document creation (\texttt{sebi:Document} \texttt{sebi:intended} \texttt{sebi:Intention}). The \texttt{dct:date} property is connected to a \texttt{time:Interval} class, which includes \texttt{time:hasBeginning} and \texttt{time:hasEnd} properties to specify the creation period and handle fuzzy time-spans. 

\begin{figure}[h]
    \centering
    \includegraphics[width=0.8\linewidth]{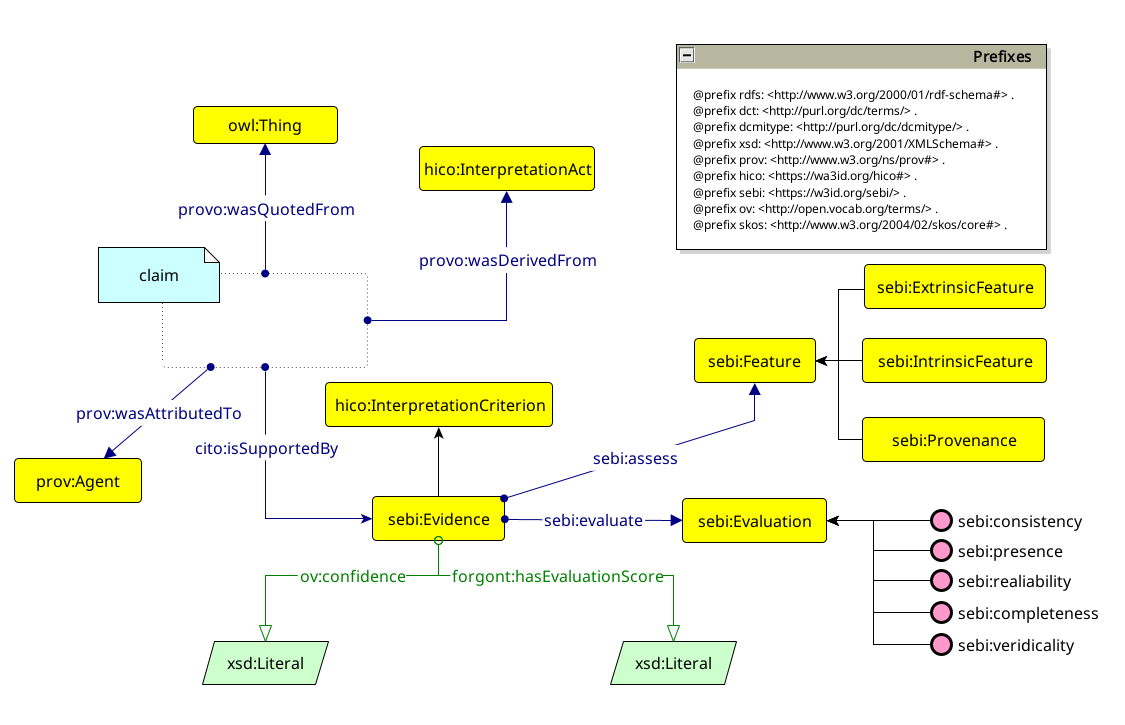}
    \caption{Selection of classes and properties to represent the contextual information about scholarly claim tackling authenticity assessment of a document}
    \label{fig:claim_datamodel}
\end{figure}

Concerning contextual information, each interpretation (set of claims represented as quoted triples) is categorised as a \texttt{hico:InterpretationAct} connected to a \texttt{prov:Agent} to address its auhtoriality and linked to the evidence supporting the claim ({\texttt{sebi:support sebi:Evidence}}). Document features and their evaluation are critical components of the ontology. Document features (\texttt{sebi:Feature}) are either extrinsic features (\texttt{sebi:ExtrinsicFeature}), intrinsic ones (\texttt{sebi:IntrinsicFeature}), or provenance information (\texttt{sebi:Provenance}), capturing aspects such as ink, support, handwriting, and orthography. Each feature is evaluated on a set of established criteria (\texttt{sebi:Evidence}) such as consistency, presence, completeness, veridicality, and reliability. A score is associated to each evidence as \texttt{xsd:Literal} using the property
\texttt{forgont:hasEvaluationScore}. The evaluation score indicates a measure on each collected evidence, allowing the integration of negatives (e.g. the absence of the signature in a document is represented as an evidence based on the feature "authentication marks" with evaluation "presence", with score \texttt{false} or $0$).

\subsection{Pilot Corpus}

Following Step II (\ref{subsec:methodology-step1}), we sampled the pilot corpus. We selected the following articles (\textit{Donation of Constantine}, \textit{Eremin Letter}, \textit{Getty Kouros}, \textit{Historia Augusta}, \textit{Life of Homer}, \textit{Marriage Charter of Empress Theophanu}, \textit{Protocols of the Elders of Sion}), each belonging to a different category. The domain expert personally pointed to the \textit{Donation of Constantine} and other medieval charters, as the \textit{Marriage Charter}, as optimal use cases for the study. 

The other articles were selected as they contained explicit scholarly disagreements about authenticity. Selection criteria included: (1) presence of multiple scholarly perspectives on authenticity, (2) clear attribution of claims to specific researchers or institutions, (3) discussion of evidence-based reasoning, and (4) representation of different temporal periods and document types.

\subsection{Minimal Working Annotation}\label{sec:annotation-model}

Our annotation model was developed through INCEpTION~\citep{inception}, following Step II of the methodology (\ref{subsec:methodology-step2}). The annotation model implements three core patterns from our ontology: \textbf{CH item metadata}, \textbf{scholarly agents}, and \textbf{authenticity opinions}.

\textbf{CH Item Metadata.} We established an identification layer using INCEpTION's Knowledge Base integration with Wikidata, allowing annotators to link textual mentions directly to Wikidata IDs for automatic coreference resolution. Item types mentioned in source texts were reconciled to DCMI Type Vocabulary classes (\texttt{dcmitype:Text}, \texttt{dcmitype:PhysicalObject}, \texttt{dcmitype:Collection}) with appropriate subclass relationships.

\textbf{Scholarly Agents.} Entities expressing opinions (Cognizers) correspond to \texttt{dct:Agent} in our ontology. Each Cognizer was linked to Wikidata when possible, with fallback strategies for entities without entries, impersonal statements, and consensus attributions.

\textbf{Authenticity Claims.} We modeled claims through directed relations between Cognizer spans and CH item spans, labeled according to SEBI's authenticity categories (\texttt{Authentic}, \texttt{FormalForgery}, \texttt{ContentForgery}, \texttt{Forgery}, \texttt{Neutral}). Each opinion becomes an RDF-star quoted triple linked to \texttt{hico:InterpretationAct}.

Using this approach on the Pilot Corpus, we validated annotation-to-RDF mapping through the algorithm in Listing~\ref{lst:mwa-mapping}, then enriched the model with additional ontology patterns.

\begin{lstlisting}[caption={Core Annotation Mapping Algorithm}, label={lst:mwa-mapping}]
STEP 1: Extract Cognizer-Opinion Pairs
Select all spans marked as Entity 
WHERE span also has Opinion tagset label
=> CognizerSet(Cognizer(CognizerSpan, Opinion, WikidataID)

STEP 2: Extract CH Items  
Select all spans marked as Entity
WHERE span has ItemTitle label
=> ItemSet(ItemSpan, WikidataID)

STEP 3: Find Relations
For CognizerSpan in CognizerSet, check if CognizerSpan
has stm:Object relation to span in ItemSet
=> Valid tuples (Cognizer, Item, Opinion)

STEP 4: Generate RDF for each tuple
For each matching pattern:
|-- Generate URI for Cognizer
|-- Add owl:sameAs + Wikidata ID
|-- Generate URI for Item  
|-- Add owl:sameAs + Wikidata ID
|-- Map opinion to corresponding SEBI class (e.g., sebi:Forgery) 
|-- Generate URI for Named Graph (hico:InterpretationAct)
|-- Generate claim triple as a RDF-star statement

+-- Apply template:
    
    ex:{cognizer_uri}_about_{item_uri} rdf:type hico:InterpretationAct ; 
    prov:wasAttributedTo ex:cognizer .

    ex:cognizer rdf:type dct:Agent ; 
    rdfs:label "CognizerSpan"@language .
    owl:sameAs wd:wikidataId
    

    ex:item rdf:type ex:type ; 
    rdfs:label "ItemSpan"@language .
    owl:sameAs wd:wikidataID

    << ex:item rdf:type sebi:Opinion >> prov:wasDerivedFrom ex:{cognizer_uri}_about_{item_uri} .
\end{lstlisting}

\begin{figure}[h!]
    \centering
    \includegraphics[width=1\linewidth]{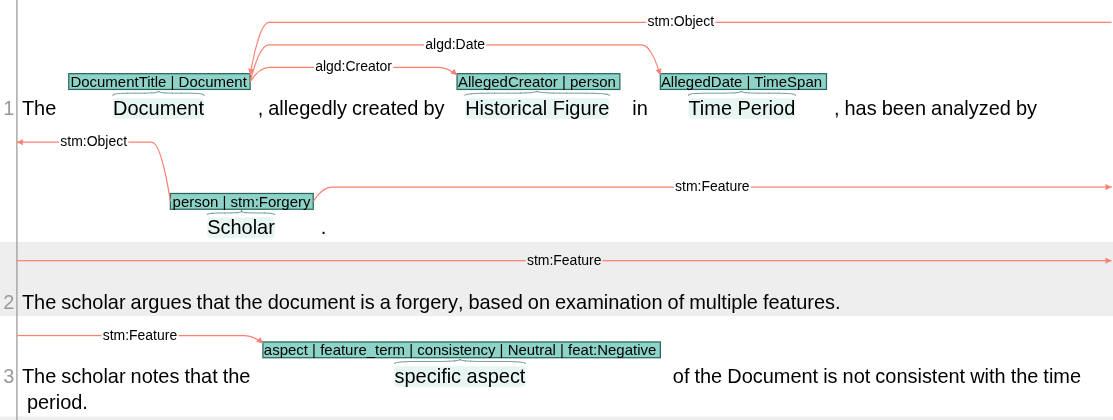}
    \caption{Example annotation of an entity expressing an opinion about a CH item}
    \label{fig:inception-entity-statement}
\end{figure}

\subsection{SEBI Annotation Model}\label{sec:final-sebi-annotation-model}
The refined annotation model captures CH item metadata, evidence and features, and scholarly hypotheses through three additional layers, satisfying Step IV of the methodology (see \ref{subsec:methodology-step4}.

CH Item Metadata Layer. This layer captures \textit{alleged metadata}—the descriptive information (creator, date, location) that the document or artifact claims about itself, representing what the item purports to be before any scholarly critical analysis. This includes the face-value claims presented by the item regarding its authorship, creation date, geographic origin, and other identifying characteristics. Annotations include \texttt{AllegedCreator}, \texttt{AllegedDate}, \texttt{AllegedLocation}, \texttt{ItemSubject}, and \texttt{ItemType}, plus properties for formal forgeries (\texttt{ItemCreator}, \texttt{ItemDate}, \texttt{ItemLocation}). 
\begin{figure}[h!]
    \centering
    \includegraphics[width=1\linewidth]{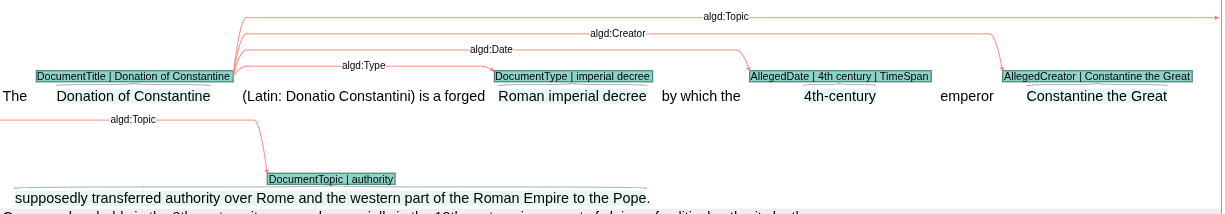}
    \caption{Alleged metadata annotation for the Donation of Constantine}
    \label{fig:annotation-donation-alleged-metadata}
\end{figure}

\textbf{Evidence and Features Layer.} This layer generates Evidence nodes connected to InterpretationAct Named Graphs. It employs four tagsets: Feature (SEBI vocabulary terms for intrinsic/extrinsic features and provenance), FeatureAssessment (evaluation perspectives: consistency, presence, completeness, reliability, veridicality), FeatureAssessmentPolarity (negative, neutral, positive), and FeatureAssessmentConfidence.

Consider Lorenzo Valla's assessment of the Donation's language features, which converts to three evidence structures linking specific textual features to evaluation criteria and polarities.

\begin{figure}[h!]
    \centering
    \includegraphics[width=1\linewidth]{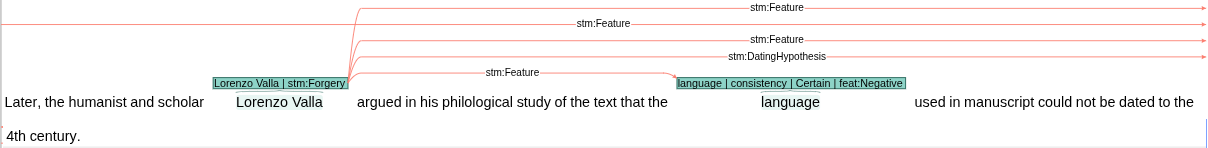}
    \caption{Lorenzo Valla's opinion with feature assessment annotation}
    \label{fig:annotation-lorenzo-valla}
\end{figure}

Listing~\ref{lst:feature-eval-mapping} shows the evidence mapping algorithm.

\begin{lstlisting}[caption={Evidence and Feature Mapping Algorithm}, label={lst:feature-eval-mapping}]
STEP 1: Extract Evaluated Features
Select all spans marked as feature
WHERE span also has FeatureAssessment label, FeatureAssessmentPolarity, FeatureAssessmentConfidence
=> FeatureSet(FeatureSpan, FeatureClass, FeatureAssessment, FeatureAssessmentPolarity, FeatureAssessmentConfidence)

STEP 2: Select all spans marked as Entity 
WHERE span also has Opinion tagset label
=> CognizerSet(Cognizer(CognizerSpan, Opinion, WikidataID)

STEP 3: Find Relations
For FeatureSpan in FeatureSet, check if CognizerSpan
has stm:Feature relation to any span(s) in FeatureSet
=> Valid tuples (Cognizer, FeatureSet)

STEP 4: Generate nodes 
For each matching pattern:
|-- Generate/Reuse URI for Cognizer
|--- Add owl:sameAs + Wikidata ID

|-- Generate URI for sebi:Evidence graph
|--- match FeatureAssessment individual with sebi:Evaluation individual 
|--- match FeatureAssessmentPolarity 
|--- attach FeatureAssessmentConfidence score

|-- Generate URI for sebi:Feature graph
|--- attach FeatureSpan through rdfs:label
|--- attach FeatureClass through skos:broader 

STEP 5: Generate RDF graph

+-- Apply template:
    
kb:{cognizer_uri}_about_{item_uri}_{idx} a sebi:Evidence ;
    sebi:assess kb:{feature_uri} ;
    sebi:evaluate sebi:{evaluation_uri} ;
    sebi:hasEvaluationScore "{polarity}"@language ;
    sebi:support kb:interpretation_act ;
    ov:confidence 1.0 .

kb:{feature_uri} a sebi:Feature ;
    rdfs:label "{FeatureSpan}"@language ;
    sebi:isAssessedBy kb:{cognizer_uri}_about_{item_uri}_{idx} ;
    skos:broader sebi:{feature_vocabulary_term} .
\end{lstlisting}
\textbf{Scholarly Hypotheses Layer.} This layer captures alternative hypotheses through four relation types linking Cognizers to Wikidata entities: \texttt{stm:CreatorHypothesis}, \texttt{stm:DatingHypothesis}, \texttt{stm:LocationHypothesis}, and \texttt{stm:ReasonHypothesis}.

\begin{figure}[h!]
    \centering
    \includegraphics[width=1\linewidth]{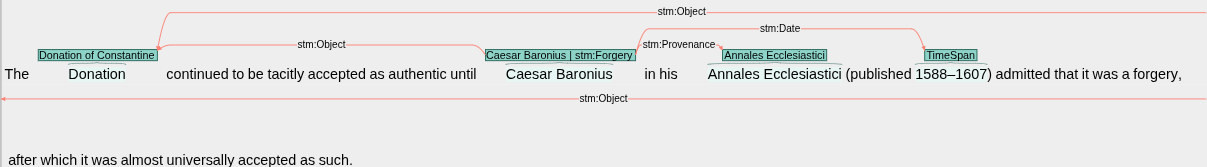}
    \caption{Caesar Baronius's admission of forgery with provenance annotation}
    \label{fig:annotation-caesar-baronius}
\end{figure}

\begin{figure}[h!]
    \centering
    \includegraphics[width=1\linewidth]{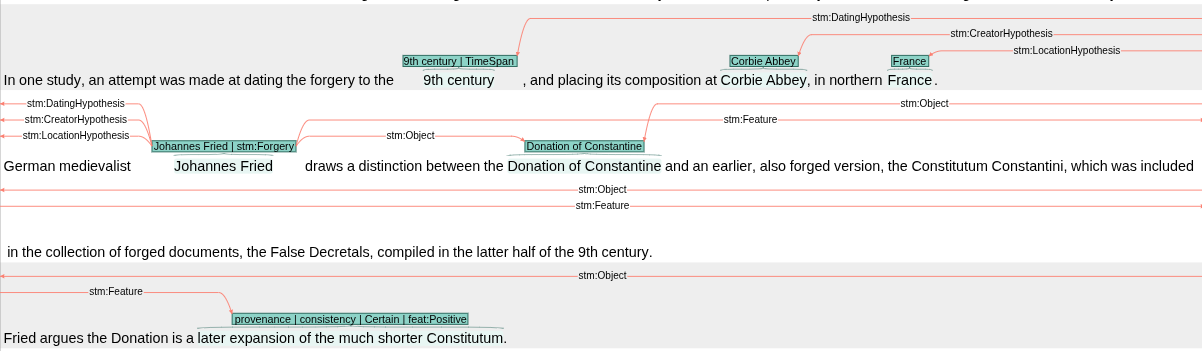}
    \caption{Johannes Fried's hypotheses annotation for the Donation of Constantine}
    \label{fig:annotation-johannes-fried}
\end{figure}

Listing~\ref{lst:hypotheses-mapping} details the hypotheses mapping algorithm.

\begin{lstlisting}[caption={Hypotheses Mapping Algorithm}, label={lst:hypotheses-mapping}]
STEP 1: Extract Hypothesis Relations
Select all relations of type:
|-- stm:CreatorHypothesis
|-- stm:DatingHypothesis  
|-- stm:LocationHypothesis
|-- stm:ReasonHypothesis
=> HypothesesSet(CognizerSpan, HypothesisType, TargetSpan, WikidataID)

STEP 2: Extract Cognizer Entities
Select all spans marked as Entity 
WHERE span also has Opinion tagset label
=> CognizerSet(CognizerSpan, Opinion, WikidataID)

STEP 3: Find Valid Patterns
For each relation in HypothesesSet:
Check if CognizerSpan exists in CognizerSet
=> Valid tuples (Cognizer, HypothesisType, Target)

STEP 4: Generate Target URIs
For each matching pattern:
|-- Generate/Reuse URI for Cognizer
|-- Generate/Reuse URI for Item
|-- Generate/Reuse URI for Target entity
|-- Map HypothesisType to corresponding RDF property

STEP 5: Generate RDF-star Statements

+-- Apply template:

kb:{target_uri} a {target_class} ;
    owl:sameAs wd:{wikidata_id} ;
    # if Wikidata ID not available
    # kb:{urifiedTargetSpan} a {target_uri} ;
    rdfs:label "{TargetSpan}"@language .

<< kb:{item_uri} dct:creator kb:{target_uri} >> 
    prov:wasDerivedFrom kb:{cognizer_uri}_about_{item_uri} .

<< kb:{item_uri} dct:date kb:{target_uri} >> 
    prov:wasDerivedFrom kb:{cognizer_uri}_about_{item_uri} .

<< kb:{item_uri} sebi:location kb:{target_uri} >> 
    prov:wasDerivedFrom kb:{cognizer_uri}_about_{item_uri} .

<< kb:{item_uri} sebi:intendedTo kb:{target_uri} >> 
    prov:wasDerivedFrom kb:{cognizer_uri}_about_{item_uri} .
\end{lstlisting}

Each annotation layer maps to RDF following SEBI ontology principles, with Wikidata integration providing entity resolution and the INCEpTION project available on GitHub alongside mapping scripts\footnote{\href{https://anonymous.4open.science/r/SEBI-Knowledge-Extraction-5FCE/README.md}{SEBI-KE repository. See 'Inception2Graph' folder}}.
Statistics for the annotation results are shown in Table \ref{tab:gt-kg-statistics}.
\begin{table}[h]
\centering
\caption{Ground Truth Annotation results}
\label{tab:gt-kg-statistics}
\begin{tabular}{lc}
\toprule
\textbf{Span} & \textbf{Count} \\
\midrule
CH Items & 45 \\
Entities & 235 \\
Interpretation Acts & 215 \\
Evidences & 132 \\
Features & 115 \\
Wikidata alignments & 308 \\
\bottomrule
\end{tabular}
\end{table}


\section{Knowledge Extraction Pipeline and Evaluation Framework}\label{sec:methodology_implementation}

This section presents the implementation of Steps III-V of our methodology (\ref{subsec:methodology-step3}, \ref{subsec:methodology-step4}, \ref{subsec:methodology-step5}), transforming the annotation model into a working KE pipeline. Our implementation integrates three complementary technologies in a sequential LLM-based process:

\begin{itemize}
    \item GliNER for lightweight NER;
    \item LLMs for structured information extraction;
    \item Rule-based entity linking for external KG integration.
\end{itemize}

Each component addresses specific challenges in CH knowledge extraction while maintaining alignment with the SEBI ontology and supporting the complex semantic dependencies characteristic of humanities discourse.

\textbf{GliNER}~\citep{zaratiana2024gliner} provides lightweight, generalist NER using custom entity types with state-of-the-art performance. 
\textbf{LLMs} handle structured information extraction through JSON schema-based responses.

\begin{figure}
    \centering
    \includegraphics[width=1\linewidth]{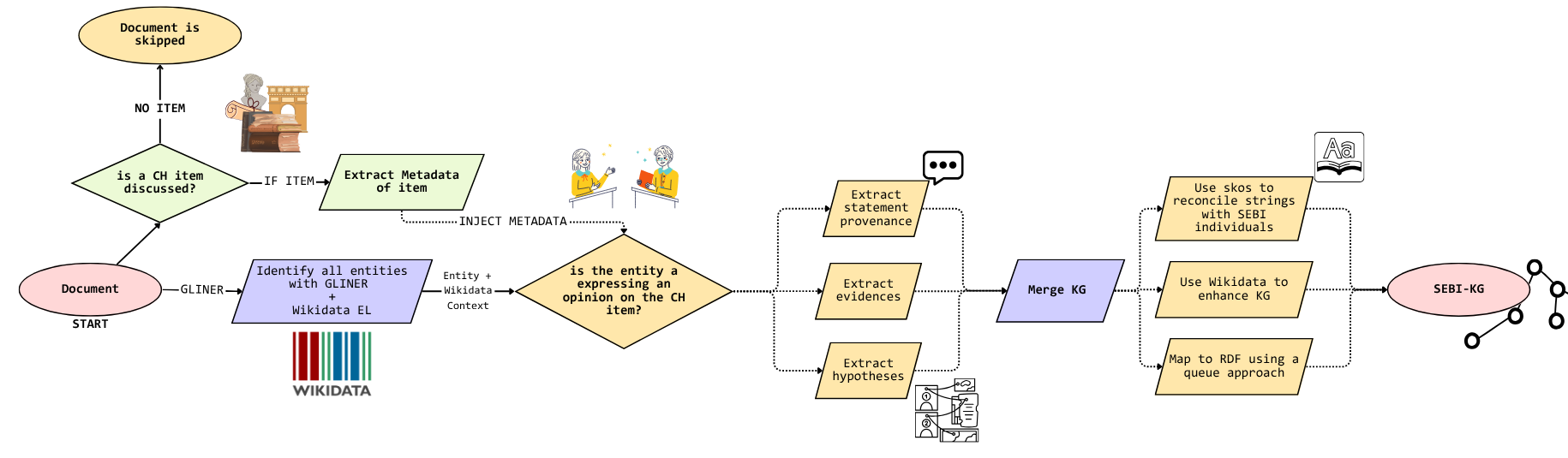}
\caption{Flowchart of the sequential pipeline for SEBI-based KG generation}\label{fig:pipeline-overview}
\end{figure}

The pipeline was used with three different models evaluated three models with different parameter scales to understand performance trade-offs: Claude Sonnet 3.7 as the biggest model\footnote{Claude Sonnet 3.7 Model Card: \url{https://assets.anthropic.com/m/785e231869ea8b3b/original/claude-3-7-sonnet-system-card.pdf}},  Llama 3.3 70B~\citep{dubey2024llama} as a medium-sized model, and GPT-4o-mini.\footnote{GPT-4o-mini model card: \url{https://openai.com/index/gpt-4o-mini-advancing-cost-efficient-intelligence/}} 
While the exact parameter size of GPT-4o-mini remains undisclosed, estimates range from 8-14 billion active parameters \citep{abacha2025medecbenchmarkmedicalerror}.
Figure \ref{fig:pipeline-overview} presents a comprehensive overview of the sequential processing pipeline.

\textbf{Entity Linking} employs a rule-based approach leveraging the Wikibase API\footnote{\href{https://www.mediawiki.org/wiki/Wikibase/API/en}{Wikibase API Documentation}} and domain-specific heuristics. After testing various state-of-the-art solutions, this approach proved most effective for historical entities and CH concepts, providing reliable external knowledge base integration while handling the specialized vocabulary of authenticity assessment debates. \\
While the selected LLMs are capable of processing documents in their entirety at any step, the system automatically selects only the relevant paragraphs whenever possible. This design serves three strategic purposes: (1) reducing content volume per processing step to minimize potential opinion overlap between entities, assuming it improves precision; (2) demonstrating the pipeline's scalability to documents of arbitrary length; and (3) maintaining computational efficiency and cost-effectiveness by minimizing token consumption per API call.

\subsection{Sequential Processing Pipeline}

The KE pipeline is based on six components, each enriching the output before passing it to the next. Each component produces a JSON, each with a given schema designed to be convertible to RDF at the end. The development of the pipeline started from previous implementations and sequential testing over the COPs until the full KG could be extracted. As the output of the JSON model is largerly similar to the output of the GT, the mapping works using a very similar logic, with the only difference of being mapped from JSON instead of the JSON UIMA CAS (Content Analysis System) format used by INCEpTION.

\begin{enumerate}
\item Raw text documents $\rightarrow$ metadata extraction $\rightarrow$ alleged/settled item metadata;
\item Item metadata + text $\rightarrow$ opinion holder identification $\rightarrow$ entity mentions with classifications;
\item Entity mentions $\rightarrow$ entity resolution $\rightarrow$ Wikidata-linked entity clusters;
\item Linked entities + paragraphs $\rightarrow$ opinion extraction $\rightarrow$ structured authenticity opinions;
\item Opinions + contexts $\rightarrow$ evidence mining $\rightarrow$ feature evaluations with polarity;
\item Evidence + full context $\rightarrow$ hypothesis extraction $\rightarrow$ conflicting statements.

\end{enumerate}

\subsubsection{CH item Metadata Extraction}\label{step:metadata-extraction}
\textbf{Aim:} Identify all the CH items being discussed in an article and get the alleged metadata about them \\
\textbf{Input:} Raw Wikipedia articles in markup (.txt files)\\
\textbf{Output:} Cleaned articles; JSON with alleged item metadata, based on a given JSON schema.

This component identifies and extracts metadata about CH items discussed in each article. 
The LLM is instructed to extract a JSON schema from the article text that describes all CH items under discussion. 
Specifically, it extracts the CH item \textit{alleged metadata}, or, in other words, what items claim to be---purported authors, creation dates, locations, item types, and subject matter.  
The task relies on In-Context Learning (ICL) in a Few-Shot setting (with 3 examples), using Chain-of-Thought (COT) reasoning. Listing \ref{lst:step-1-input-putput} shows an example text extracted from the Donation of Constantine Wikipedia article as input (on the left) and the JSON output (on the right).

\renewcommand{\figurename}{Listing}
\begin{figure}[h]
\begin{minipage}[t]{0.48\textwidth}
\begin{tcolorbox}[
  title={Input source text},
  fonttitle=\bfseries,
  colframe=customred!70!black,
  colback=customred!5!white,
  width=\textwidth,
  top=2mm,
  bottom=2mm,
  left=2mm,
  right=2mm
]
The \textbf{Donation of Constantine} [...] is a forged Roman imperial decree by which the 4th-century emperor Constantine the Great supposedly transferred authority over Rome and the western part of the Roman Empire to the Pope [...]. [I]t was used, especially in the 13th century, in support of claims of political authority by the papacy.

\end{tcolorbox}
\end{minipage}
\hfill
\begin{minipage}[t]{0.48\textwidth}
\begin{tcolorbox}[
  title={JSON output},
  fonttitle=\bfseries,
  colframe=customgreen!70!black,
  colback=customgreen!5!white,
  listing only,
  width=\textwidth,
  top=2mm,
  bottom=2mm,
  left=2mm,
  right=2mm,
  listing options={basicstyle=\tiny\fontfamily{pcr}\selectfont}
]
\texttt{\{\\
~~"item": "Donation of Constantine",\\
~~~"alleged\_author": "Constantine the Great",\\
~~~"alleged\_date": "4th century",\\
~~~"alleged\_location": "Rome",\\
~~~"item\_type": "decree",\\
\}}
\vfill
\end{tcolorbox}
\end{minipage}
\caption{CH Item metadata extraction from source text to structured JSON output}
\label{lst:step-1-input-putput}
\end{figure}

\subsubsection{Cognizer Identification}\label{step:cognizer-identification}
\textbf{Aim:} Identify the subset of Cognizers who make scholarly statements around the given CH items \\
\textbf{Input:} Cleaned article + item(s) metadata (output of previous step, see Section \ref{step:metadata-extraction})\\
\textbf{Output:} JSON with \texttt{is\_cognizer} classification, coreferences 

This component employs GliNER \citep{zaratiana2024gliner} for NER, targeting people, organizations, groups, and locations. GliNER identifies precise character-level spans compared to LLMs, enabling us to exactly identify and group the paragraphs in which each entity appears.
Among the selected paragraphs, we aimed at selecting exclusively the subset of entities who actually express opinions, omitting those who do not from later steps. We created a prompt which instructs the model to check if the entity extracted from GliNER is expressing an opinion on the CH item(s) extracted in Step 1 (\ref{step:metadata-extraction}). The prompt asks to return a binary classification (is\_expressing\_opinion: True/False) alongside additional textual mentions and co-references of the given entity. The task relies on In-Context Learning (ICL) in a Few-Shot setting (with 3 examples), using Chain-of-Thought (COT) reasoning.
Listing \ref{lst:step-2-input-putput} shows a paragraph where \textbf{Lorenzo Valla } is mentioned alongside the output.

\renewcommand{\figurename}{Listing}
\begin{figure}[h]
\begin{minipage}[t]{0.48\textwidth}
\begin{tcolorbox}[
  title={Input: Source text},
  fonttitle=\bfseries,
  colframe=customred!70!black,
  colback=customred!5!white,
  width=\textwidth,
  top=2mm,
  bottom=2mm,
  left=2mm,
  right=2mm
]
Later, the humanist and scholar \textbf{Lorenzo Valla} argued in his philological study of the text that the language used in manuscript could not be dated to the 4th century.[21] The language of the text suggests that the manuscript can most likely be dated to the 8th century. \textbf{Valla} believed the forgery to be so obvious that he suspected that the Church knew the document to be inauthentic. 

\end{tcolorbox}
\end{minipage}
\hfill
\begin{minipage}[t]{0.48\textwidth}
\begin{tcolorbox}[
  title={Input: JSON output},
  fonttitle=\bfseries,
  colframe=customgreen!70!black,
  colback=customgreen!5!white,
  listing only,
  width=\textwidth,
  top=2mm,
  bottom=2mm,
  left=2mm,
  right=2mm,
  listing options={basicstyle=\tiny\fontfamily{pcr}\selectfont}
]
\texttt{\{\\
~~"entity": "Lorenzo Valla",\\
~~"start": 605,\\
~~"end": 618,\\
~~"label": "person",\\
~~"is\_cognizer": true,\\
~~"is\_subject": true,\\
~~"mentions": [\\
~~~~"Lorenzo Valla",\\
~~~~"Valla"\\
]\\
\}}
\vfill
\end{tcolorbox}
\end{minipage}
\caption{CH Item extraction from source text to structured JSON output}
\label{lst:step-2-input-putput}
\end{figure}


\subsubsection{Entity Resolution and Linking}
\textbf{Aim:} Enrich Cognizers with biographical information; paragraphs grouped by Cognizer\\
\textbf{Input:} Cognizers and coreferences\\
\textbf{Output:} JSON with relevant paragraphs grouped by Cognizer and its biographical information\\

The third component performs coreference resolution and Entity Linking (EL). It clusters entities through identical mentions across paragraphs, then uses the Wikibase API\footnote{\url{https://www.mediawiki.org/wiki/Wikibase/API}} to retrieve candidates for the longest mention of each entity. Each candidate receives a score based on name similarity (Levenshtein distance\footnote{\url{https://pypi.org/project/python-Levenshtein/}}) between mentions and Wikidata labels/aliases, entity type compatibility, and for people, occupation relevance using Wikidata property \texttt{P106}\footnote{\url{https://www.wikidata.org/wiki/Property:P106}} (prioritizing scholarly occupations likely to express opinions in this domain).
\renewcommand{\figurename}{Listing}
\begin{figure}[h]
\begin{minipage}[t]{0.48\textwidth}
\begin{tcolorbox}[
  title={Input: Clustered entities},
  fonttitle=\bfseries,
  colframe=customred!70!black,
  colback=customred!5!white,
  listing only,
  width=\textwidth,
  top=2mm,
  bottom=2mm,
  left=2mm,
  right=2mm,
  listing options={basicstyle=\tiny\fontfamily{pcr}\selectfont}
]
\texttt{\{\\
~~~~"primary\_mention": "Lorenzo Valla",\\
~~~~"all\_mentions": [\\
~~~~~~"Lorenzo Valla",\\
~~~~~~"Valla",\\
~~~~~~"the humanist"\\
~~~~],\\
~~~~"entity\_type": "person",\\
~~~~"paragraphs": [0, 3, 7]\\
\}}
\end{tcolorbox}
\end{minipage}
\hfill
\begin{minipage}[t]{0.48\textwidth}
\begin{tcolorbox}[
  title={Output: Wikidata entity},
  fonttitle=\bfseries,
  colframe=customgreen!70!black,
  colback=customgreen!5!white,
  listing only,
  width=\textwidth,
  top=2mm,
  bottom=2mm,
  left=2mm,
  right=2mm,
  listing options={basicstyle=\tiny\fontfamily{pcr}\selectfont}
]
\texttt{\{\\
~~"wikidata\_label": "Lorenzo Valla",\\
~~"wikidata\_id": "Q214115",\\
~~"occupation": ["humanist", "philologist"...],\\
~~"birth\_year": 1407,\\
~~"death\_year": 1457,\\
~~"mentions": ["Lorenzo Valla", "Valla"]\\
\}}
\end{tcolorbox}
\end{minipage}
\caption{Entity resolution and linking: from clustered mentions to Wikidata-enriched entities}
\label{lst:step-3-input-output}
\end{figure}

\subsubsection{Opinion Extraction and Classification}\label{step:opinion-extraction}
\textbf{Aim:} Extract the first layer of the Cognizer's opinion \\
\textbf{Input:} Entity + Wikidata Information (if linked) + paragraphs where entity is mentioned\\
\textbf{Output:} JSON describing (1) the Cognizer's opinion, (2) their opinion, the metadata of the opinion (where, when, provenance) 

The fourth component extracts and classifies authenticity opinions based on \ref{step:cognizer-identification}. If the entity has been successfully linked to Wikidata, this information is given to the model as well. 

The extraction process captures the main COP of the opinion: the opinion target(s) (which documents or artifacts), opinion types following SEBI classifications (Authentic, Forgery, Formal forgery, Content forgery, Neutral), confidence levels expressed by the Cognizer, temporal contexts (when opinions were expressed), and geographic contexts where relevant. See Listing \ref{lst:step-4-opinion-output} as reference.

\renewcommand{\figurename}{Listing}
\begin{figure}[h]
\begin{minipage}[t]{0.48\textwidth}
\begin{tcolorbox}[
  title={Input: Source text},
  fonttitle=\bfseries,
  colframe=customred!70!black,
  colback=customred!5!white,
  width=\textwidth,
  top=1mm,
  bottom=1mm,
  left=1mm,
  right=1mm
]
{Later, the humanist and scholar \textbf{Lorenzo Valla} argued in his philological study of the text that the language used in \textbf{the manuscript} could not be dated to the 4th century. The language of the text suggests that the manuscript can most likely be dated to the 8th century. Valla believed the \textbf{forgery} to be so obvious that he suspected that the Church knew the document [...]}
\end{tcolorbox}
\end{minipage}
\hfill
\begin{minipage}[t]{0.48\textwidth}
\begin{tcolorbox}[
  title={JSON output: Opinion classification},
  fonttitle=\bfseries,
  colframe=customgreen!70!black,
  colback=customgreen!5!white,
  listing only,
  width=\textwidth,
  top=1mm,
  bottom=1mm,
  left=1mm,
  right=1mm,
  listing options={basicstyle=\scriptsize\fontfamily{pcr}\selectfont}
]
\texttt{"opinions":}\\
\texttt{[\{}\\
\texttt{~~"entity": "Lorenzo Valla",}\\
\texttt{~~"subject": "Donation of Constantine",}\\
\texttt{~~"opinion": "Forgery",}\\
\texttt{~~"confidence": "High",}\\
\texttt{~~"date": "1439-1440"}\\
\texttt{~~"location": ""}\\
\texttt{\}]}
\end{tcolorbox}
\end{minipage}
\caption{JSON output from source text about Cognizer, subject of the opinion, provenance, assessment}
\label{lst:step-4-opinion-output}
\end{figure}

\subsubsection{Evidence Mining and Feature Assessment}\label{step:evidence-mining}
\textbf{Aim:} Enrich the Cognizer's opinion with the supporting evidence \\
\textbf{Input:} Structured opinions + contextual paragraphs\\
\textbf{Output:} JSON with supporting evidences and evaluations for each opinion

In the fifth component, the model is tasked with enriching the basic opinion of the Cognizer with evidences and features being evaluated

Features are organized into three categories following the SEBI ontology: \textit{intrinsic features} (content, language, style, orthography), \textit{extrinsic features} (handwriting, ink, material support, physical characteristics), and \textit{provenance information} (historical context, witness accounts, transmission history).
For each feature, the system determines evaluation criteria including consistency (does the feature match the alleged period/author?), presence (is the expected feature present or absent?), completeness (is the feature fully preserved/documented?), reliability (can the feature be trusted as evidence?), and veridicality (does the feature represent authentic information?).

Each evaluation receives polarity assignment (positive, negative, neutral evidence) and links to supporting scholarly opinions, creating structured representations of evidence-based reasoning in authenticity assessment. See Listing \ref{lst:step-5-input-output} as reference.

\renewcommand{\figurename}{Listing}
\begin{figure}[h]
\begin{minipage}[t]{0.48\textwidth}
\begin{tcolorbox}[
  title={Input: Source text},
  fonttitle=\bfseries,
  colframe=customred!70!black,
  colback=customred!5!white,
  width=\textwidth,
  top=1mm,
  bottom=1mm,
  left=1mm,
  right=1mm
]
{[...] \textbf{Reginald Pecocke}, Bishop of Chichester (1450–57), reached a similar conclusion. Among the indications that the Donation is a forgery are its language and the fact that, while certain \textbf{imperial-era formulas} are \textbf{used in the text}, some of the Latin in the document could not have been written in the 4th century}
\end{tcolorbox}
\end{minipage}
\hfill
\begin{minipage}[t]{0.48\textwidth}
\begin{tcolorbox}[
  title={JSON output: Evidence extraction},
  fonttitle=\bfseries,
  colframe=customgreen!70!black,
  colback=customgreen!5!white,
  listing only,
  width=\textwidth,
  top=1mm,
  bottom=1mm,
  left=1mm,
  right=1mm,
  listing options={basicstyle=\scriptsize\fontfamily{pcr}\selectfont}
]
\begin{verbatim}
"evidence_evaluations":
[{
  "evidence": 
  "imperial-era formulas",
  "feature": "language", 
  "evaluation": "presence",
  "polarity": "positive"
}]
\end{verbatim}
\end{tcolorbox}
\end{minipage}
\caption{JSON output of the identified evidences with evaluation}
\label{lst:step-5-input-output}
\end{figure}

\subsubsection{Hypothesis Extraction}\label{step:hypotheses-extraction}
\textbf{Aim:} Enrich the Cognizer's opinion with hypotheses made on the CH item(s)  \\
\textbf{Input:} Opinions + evidence evaluations + full document context\\
\textbf{Output:} JSON with hypotheses about document origins, intent, etc.

The final component enriches the output of the previous one with the Cognizer's hypotheses on the CH item. The hypotheses can be of four types: \textit{authorship hypotheses} (who actually created items if not alleged authors?), \textit{dating hypotheses} (when were items actually created if not alleged dates?), \textit{location hypotheses} (where were items actually created if not alleged locations?), and \textit{motivation hypotheses} (why were items created or forged?).

The system handles cases where Cognizers accept alleged metadata as authentic as well. For consistency and to avoid negated categories (e.g. "not Constantine", we include polarity (positive/negative) as a field. See Listing \ref{lst:step-6-hypothesis-output} as reference.
\renewcommand{\figurename}{Listing}
\begin{figure}[h]
\begin{minipage}[t]{0.48\textwidth}
\begin{tcolorbox}[
  title={Input: Source text},
  fonttitle=\bfseries,
  colframe=customred!70!black,
  colback=customred!5!white,
  width=\textwidth,
  top=1mm,
  bottom=1mm,
  left=1mm,
  right=1mm
]
{\small Later, the humanist and scholar \textbf{Lorenzo Valla} argued in his philological study of the text that the language used in manuscript could not be dated to the 4th century. The language of the text suggests that the manuscript can most likely be dated to the \textbf{8th century} [...] Valla further argued that \textbf{papal usurpation of temporal power} had corrupted the church, caused the wars of Italy, and reinforced the "overbearing, barbarous, tyrannical priestly domination."}
\end{tcolorbox}
\end{minipage}
\hfill
\begin{minipage}[t]{0.48\textwidth}
\begin{tcolorbox}[
  title={JSON output: Hypothesis extraction},
  fonttitle=\bfseries,
  colframe=customgreen!70!black,
  colback=customgreen!5!white,
  listing only,
  width=\textwidth,
  top=1mm,
  bottom=1mm,
  left=1mm,
  right=1mm,
  listing options={basicstyle=\scriptsize\fontfamily{pcr}\selectfont}
]
\texttt{"hypotheses":}\\
\texttt{\{}\\
\texttt{~~"authorship": \{}\\
\texttt{~~~~"hypothesis": "Constantine",}\\
\texttt{~~~~"confidence": "High"}\\
\texttt{~~~~"polarity": "negative"}\\
\texttt{~~\},}\\
\texttt{~~"creation\_date": \{}\\
\texttt{~~~~"hypothesis": "8th century",}\\
\texttt{~~~~"confidence": "Medium"}\\
\texttt{~~~~"polarity": "positive"}\\
\texttt{~~\}}
\end{tcolorbox}
\end{minipage}
\caption{Hypothesis extraction from source text to structured alternative theories}
\label{lst:step-6-hypothesis-output}
\end{figure}

\subsection{Knowledge Graph}
The final output is then mapped to RDF using the algorithms explained in Sections \ref{sec:annotation-model} and  \ref{sec:final-sebi-annotation-model}. This subsection serves as a showcase of the produced KGs in RDF-star and to describe the anatomy of our outputs (specifically, this was generated by the pipeline using Llama 3.3 70B).
\renewcommand{\figurename}{Figure}
\begin{figure}
    \centering
    \includegraphics[width=1\linewidth]{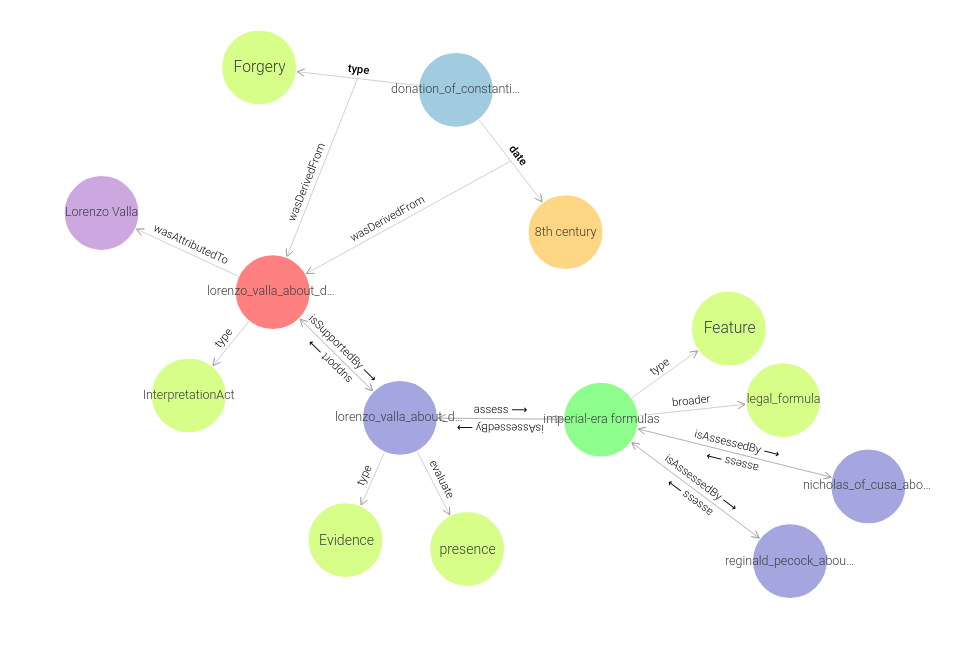}
    \caption{Lorenzo Valla's statement about the Donation of Constantine}
    \label{fig:kg-structure}
\end{figure}
Figure~\ref{fig:kg-structure} shows the general structure of a generated KG from the GraphDB interface \citep{ontotext_graphdb_2024}. Each CH item is represented with both alleged metadata (what the item(s) claim(s) to be) and scholarly assessments, as shown in Listing~\ref{lst:document-representation}. The Donation of Constantine exemplifies this pattern:

\begin{lstlisting}[caption={Document representation with alleged and scholarly metadata}, label={lst:document-representation}, language=SQL]
# Basic Item information
kb:donation_of_constantine a sebi:Decree ;
    dct:title "Donation of Constantine"@en ;
    dct:coverage kb:rome .

# Item type definition, generated from the text:
sebi:Decree rdfs:subClassOf dcmitype:Text ;
    rdfs:label "decree"@en .

# Alleged metadata as quoted triples (what the item purports to be)
<< kb:donation_of_constantine dct:creator kb:constantine_the_great >> 
    prov:wasDerivedFrom kb:donation_of_constantine_self_statement .

<< kb:donation_of_constantine dct:date kb:march_30_no_year_specified_but_implied_to_be_during_constantines_reign_306-337_ad >> 
    prov:wasDerivedFrom kb:donation_of_constantine_self_statement .

<< kb:donation_of_constantine dct:coverage kb:rome >> 
    prov:wasDerivedFrom kb:donation_of_constantine_self_statement .
\end{lstlisting}

Listing \ref{lst:valla-interpretation} shows Lorenzo Valla's interpretation of the Donation. 

\begin{lstlisting}[caption={Lorenzo Valla's interpretation with supporting evidence}, label={lst:valla-interpretation},language=SQL, showstringspaces=false]
# Lorenzo Valla as scholarly agent
kb:lorenzo_valla a sebi:Human, dct:Agent ;
    rdfs:label "Lorenzo Valla"@en ;
    owl:sameAs wd:Q214115 ;
    skos:altLabel "Valla"@en ;
    wd:occupation kb:latin_catholic_priest, kb:philologist, 
                 kb:philosopher, kb:renaissance_humanist .

# Valla's interpretation act
kb:lorenzo_valla_about_donation_of_constantine a hico:InterpretationAct ;
    sebi:date kb:1439-1440 ;
    prov:wasAttributedTo kb:lorenzo_valla ;
    prov:wasQuotedFrom "donation_of_constantine"^^xsd:anyURI ;
    cito:isSupportedBy kb:lorenzo_valla_about_donation_of_constantine_1 .

# Main authenticity claim
<< kb:donation_of_constantine rdf:type sebi:Forgery >> 
    prov:wasDerivedFrom kb:lorenzo_valla_about_donation_of_constantine .

# Alternative dating hypothesis
<< kb:donation_of_constantine dct:date kb:8th_century >> 
    prov:wasDerivedFrom kb:lorenzo_valla_about_donation_of_constantine .

# Motivation hypothesis
<< kb:donation_of_constantine sebi:intendedTo kb:political_authority >> 
    prov:wasDerivedFrom kb:lorenzo_valla_about_donation_of_constantine .
\end{lstlisting}

The supporting evidence for Valla's conclusions is captured through the Evidence graph, shown in Listing \ref{lst:valla-evidence}.

\begin{lstlisting}[caption={Lorenzo Valla's philological evidence structure}, label={lst:valla-evidence}, language=SQL]
# Evidence node linking feature assessment to interpretation
kb:lorenzo_valla_about_donation_of_constantine_1 a sebi:Evidence ;
    sebi:assess kb:philological_arguments ;
    sebi:evaluate sebi:consistency ;
    sebi:hasEvaluationScore "negative"@en ;
    sebi:support kb:lorenzo_valla_about_donation_of_constantine ;
    ov:confidence 1.0 .

# Feature being assessed
kb:philological_arguments a sebi:Feature ;
    rdfs:label "philological arguments"@en ;
    sebi:isAssessedBy kb:lorenzo_valla_about_donation_of_constantine_1 ;
    skos:broader kb:language .
\end{lstlisting}

\subsection{Evaluation Framework}
\label{subsec:evaluation-framework}

Our evaluation framework provides a multi-dimensional assessment of the KG generation pipeline, as described in Section~\ref{subsec:methodology-step5}, evaluating both the automated extraction components and the overall discourse representation quality. We integrate human assessment throughout our evaluation pipeline to align KGs, and use F$_1$ score and G-EVAL.
The framework was implemented to systematically addresses five Evaluation Questions (EQs), in line with RQ2 to RQ5 (See \ref{research-questions}).

\textbf{EQ1: CH Item Metadata Extraction Precision:} How accurately does the pipeline extract alleged item metadata compared to expert annotations?

\textbf{Methodology:} We formulate this as a multiclass classification task, evaluating the metadata extraction component described in Sec~\ref{step:metadata-extraction} against the GT. Our classification scheme follows standard evaluation practices:

\begin{itemize}
    \item \textbf{True Positive (TP):} Exact matches between model output and GT
    \item \textbf{False Positive (FP):} Incorrect model predictions
    \item \textbf{True Negative (TN):} Correctly identified absence of metadata when GT is also empty
    \item \textbf{False Negative (FN):} Missing outputs when GT contains valid metadata
\end{itemize}

To accommodate acceptable semantic variations (e.g., alternative titles, location aliases), we manually review all FP cases to identify outputs that are semantically equivalent to the GT and should be reclassified as TP.

\textbf{Metrics:} We report micro-averaged results for individual metadata categories (Title, Creator, Date, Location) and macro-averaged overall performance using standard precision, recall, and F$_1$-score calculations.\\

\textbf{EQ2: Scholarly Entity Recognition Coverage} How effectively does the entity recognition and opinion frame module identify scholarly agents (Cognizers) present in the source documents?

\textbf{Methodology:} We evaluate the entity extraction component by conducting frequency-based analysis comparing GT entities with model-identified entities as described in Section~\ref{step:cognizer-identification}.

\textbf{Metrics:} We calculate entity-level recall (proportion of GT entities correctly identified) and report the total number of entities detected by the model to assess both coverage and potential over-generation.\\

\textbf{EQ3: Evidential Reasoning Extraction Quality} How accurately does the model capture the multi-dimensional evidential reasoning employed by scholars in their interpretations?

\textbf{Methodology:} Given the complex structure of scholarly evidence identified in our ontological framework (Section~\ref{subsec:ontology}), where each piece of evidence comprises multiple semantic dimensions (evaluated feature, evaluation perspective, broader feature class, polarity), we implement a custom scoring metric operating on a 4-point scale.

For each evidence prediction, we assign points based on accuracy across these four dimensions, subtracting one point for each incorrectly identified component. This approach accommodates cases where model outputs are semantically similar but not lexically identical to GT annotations.

\textbf{Example:} Consider a scholar arguing that a document is forged due to linguistic anachronisms. If the GT annotation records ``lack of regional terms - language - presence - negative'' but the model outputs ``expected language variety - language - consistency - negative,'' this represents acceptable semantic alignment despite surface-level differences, warranting partial credit rather than complete penalization.

\textbf{Score Interpretation:}
\begin{itemize}
    \item \textbf{0 points:} Complete extraction failure (equivalent to FN or total FP)
    \item \textbf{1-2 points:} Weak but partially acceptable outputs
    \item \textbf{3-4 points:} Acceptable to strong outputs meeting semantic requirements
\end{itemize}

\textbf{Scope:} Evidence evaluation is restricted to entities successfully matched between model output and GT from RQ2.\\

\textbf{EQ4: Hypothesis and Judgment Identification} How accurately does the model extract scholars' interpretative hypotheses and overall authenticity judgments?

\textbf{Methodology:} We apply the same precision, recall, and F$_1$-score evaluation framework established for RQ1 to assess the hypothesis extraction component described in Section~\ref{step:hypotheses-extraction}. Model outputs are compared against expert-annotated GT for both specific scholarly hypotheses and overall authenticity determinations.

\textbf{Scope:} Evaluation is limited to the subset of successfully matched entities identified in RQ2 to ensure fair comparison.\\

\textbf{EQ5: Overall Discourse Representation Fidelity} Does the complete generated KG provide an adequate representation of the scholarly debate surrounding the CH items' authenticity?

\textbf{Methodology} To evaluate the fidelity of the representation we decided to use G-EVAL \citep{geval-evaluation}. Since the KGs only represent the opinions inside the text, comparing the source document with a rehydrated version of the KG would heavily bias the evaluation metric. This led us to avoid similarity-based metrics like BLEU, ROUGE and COMET with the source corpus as in \citep{gangemi_musicbo_2024}.
We chose G-EVAL to evaluate two metrics: \textit{debate correctness} and \textit{debate representativeness}. The first evaluates how well individual scholarly entities and their arguments are represented compared to the GT, penalizing omission of specific entities while rewarding accurate representation of facts, claims, and evidence with proper domain-specific terminology. The second assesses how comprehensively the overall structure and flow of the authenticity debate is captured, including the breadth of scholarly perspectives and their relationships within the discourse narrative.
It must be taken into account that previous evaluation mostly covered matchable entries between GT and output. G-EVAL evaluates the whole output. 

\textbf{Scope:} G-EVAL over rehydrated KGs covers the whole pipeline output against the rehydrated GT. 

\section{Results}\label{sec:results}
This section presents a comprehensive evaluation and preliminary discussion of findings across the five evaluation questions (EQs) outlined in Section~\ref{subsec:evaluation-framework}. We evaluate Claude Sonnet 3.7, Llama 3.3 70B, and GPT-4o-mini across multiple dimensions of the authenticity debate extraction task (the tables will show only Claude, GPT, Llama for brevity). We begin with simple exploratory SPARQL queries across the 3 KGs and compare the results with the GT, as shown in Listing \ref{lst:sparql-entity-count}. 

\renewcommand{\figurename}{Listing}
\begin{figure}[h]
\begin{tcolorbox}[
  title={SPARQL Query for KG Statistics},
  fonttitle=\bfseries,
  colframe=blue!70!black,
  colback=blue!5!white,
  width=\textwidth,
  top=2mm,
  bottom=2mm,
  left=2mm,
  right=2mm
]
\begin{lstlisting}[language=SQL, morekeywords={PREFIX, GRAPH, FILTER, CONTAINS, STR, hico, prov, dct, InterpretationAct, wasAttributedTo, Agent}, basicstyle=\small\fontfamily{pcr}\selectfont, breaklines=true]
SELECT (COUNT(DISTINCT ?entity) AS ?entityCount)
WHERE {
  ?interpretationAct a hico:InterpretationAct .
  ?interpretationAct prov:wasAttributedTo ?entity .
  ?entity a dct:Agent .
  
  FILTER(!CONTAINS(STR(?interpretationAct), "self_statement"))
}
\end{lstlisting}
\end{tcolorbox}
\caption{SPARQL query used to extract entity counts from the KGs for statistical comparison across models}
\label{lst:sparql-entity-count}
\end{figure}

\begin{table}[h]
\centering
\caption{KE overall metrics}
\label{tab:extraction-results}
\begin{tabular}{lrcc}
\toprule
\textbf{Model} & \textbf{Triples} & \textbf{Interpretation Acts} & \textbf{Cognizers} \\
\midrule
Ground Truth & 4,026 & 170 & 164  \\
Claude & 10,173 & 148 & 103\\
GPT & 12,088 & 247 & 201\\
Llama & 10,119 & 217 & 172\\
\bottomrule
\end{tabular}
\end{table}

\begin{figure}
    \centering
    \includegraphics[width=0.8\linewidth]{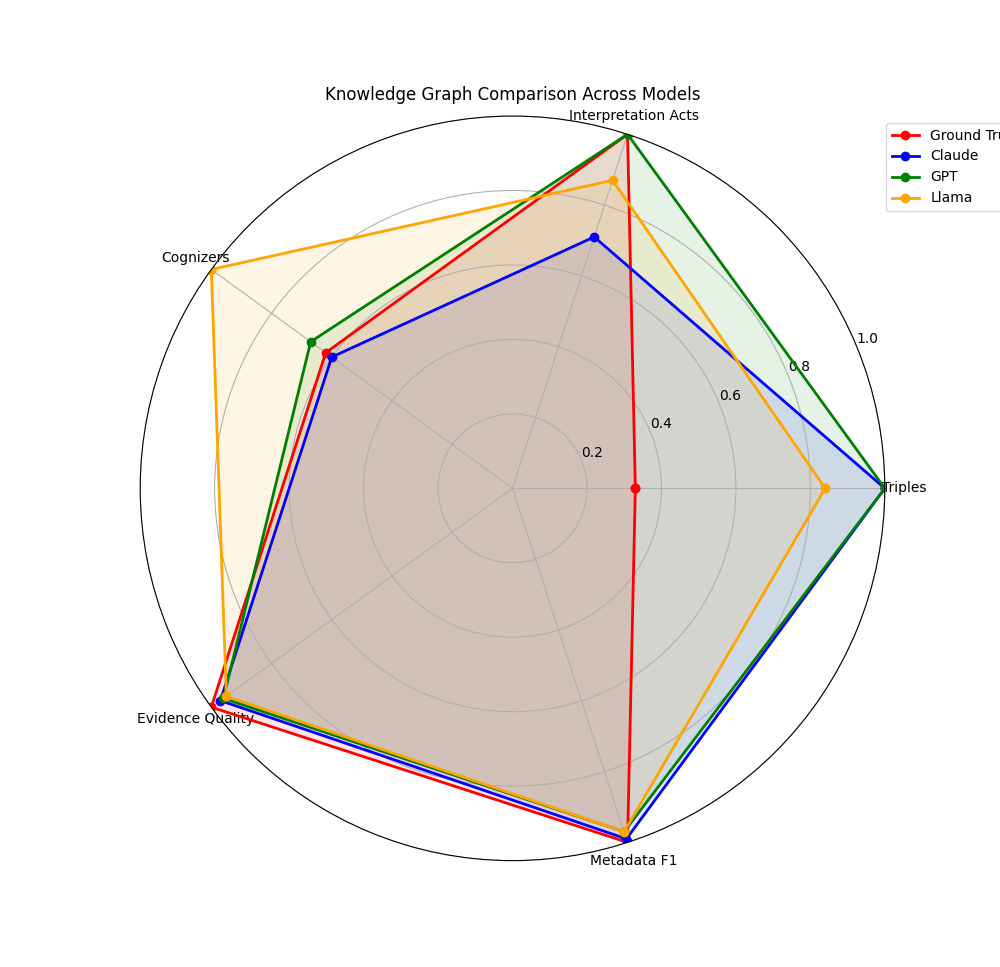}
    \caption{Radar Chart of different KG extractions}
    \label{fig:radar-chart-general}
\end{figure}

Table~\ref{tab:extraction-results} and Image \ref{fig:radar-chart-general} provide an overview of the KGs generated by each model compared to the GT. The models produces more triples than the GT (10,000-12,000 vs. 4,026), primarily because the GT relies heavily on Wikidata entity linking, while the models extract and create explicit triples for information found directly in the text (such as dates, locations, and descriptive metadata). Despite this difference in triple count, the models generate comparable numbers of Interpretation Acts and Cognizers to the GT, suggesting at this stage a similar density of extracted information.

\subsection{EQ1: CH Item Metadata Extraction Precision}\label{subsec:eq1}

\textbf{How accurately does the pipeline extract alleged CH Item metadata compared to expert annotations?}

As shown in Table~\ref{tab:metadata-results}, the performance is high across all models, with F$_1$-scores ranging from 0.97 to 0.99. 

\begin{table}[h]
\centering
\caption{CH Item Metadata extraction performance across three LLMs}
\label{tab:metadata-results}
\begin{tabular}{llccc}
\toprule
\textbf{Model} & \textbf{Category} & \textbf{Precision} & \textbf{Recall} & \textbf{F$_1$-Score} \\
\midrule
\multirow{6}{*}{\textbf{Claude}} & Titles & 1.000 & 1.000 & 1.000 \\
 & Type & 1.000 & 1.000 & 1.000 \\
 & Creators & 0.977 & 0.977 & 0.977 \\
 & Dates & 0.978 & 1.000 & 0.989 \\
 & Locations & 0.978 & 1.000 & 0.989 \\
 & \textbf{Overall} & \textbf{0.987} & \textbf{0.995} & \textbf{0.991} \\
\midrule
\multirow{6}{*}{\textbf{GPT}} & Titles & 0.889 & 1.000 & 0.941 \\
 & Type & 0.956 & 1.000 & 0.977 \\
 & Creators & 0.956 & 1.000 & 0.977 \\
 & Dates & 0.911 & 1.000 & 0.953 \\
 & Locations & 1.000 & 1.000 & 1.000 \\
 & \textbf{Overall} & \textbf{0.942} & \textbf{1.000} & \textbf{0.970} \\
\midrule
\multirow{6}{*}{\textbf{Llama}} & Titles & 0.933 & 1.000 & 0.966 \\
 & Type & 0.933 & 1.000 & 0.966 \\
 & Creators & 0.933 & 1.000 & 0.966 \\
 & Dates & 0.867 & 1.000 & 0.929 \\
 & Locations & 1.000 & 1.000 & 1.000 \\
 & \textbf{Overall} & \textbf{0.933} & \textbf{1.000} & \textbf{0.965} \\
\bottomrule
\end{tabular}
\end{table}
Claude Sonnet 3.7 achieves the highest overall performance with an F$_1$-score of 0.987. All models show nearly perfect recall, indicating successful extraction of all relevant metadata elements, with precision differences primarily reflecting varying false positive rates. Date extraction shows more variability, with Llama 3.3 achieving the lowest precision (0.867) due to higher FPs rates, as it misclassified the forging date with the alleged dating. For this particular task the challenge was to distinguish between alleged metadata and settled metadata. All models successfully understood the task, showing only small precision drops at varying parameter size.

\subsection{EQ2: Scholarly Entity Recognition Coverage}\label{subsec:eq2}

\textbf{How effectively does the entity recognition and opinion frame module identify scholarly agents (Cognizers) present in the source documents?}
As shown in Table~\ref{tab:extraction-results}, the number of Cognizers is relatively similar across models - Table~\ref{tab:entity-coverage} shows the number of overlapping entities between the model's KG and the GT. 

\begin{table}[h]
\centering
\caption{Entity recognition coverage and accuracy}
\label{tab:entity-coverage}
\begin{tabular}{lcccrcc}
\toprule
\textbf{Model} & \textbf{Precision} & \textbf{Recall} & \textbf{F$_1$} & \textbf{TP} & \textbf{FP} & \textbf{FN}\\
\midrule
\textbf{Claude} & 0.696 & 0.763 & 0.728 & 71 & 31 & 22\\
\textbf{GPT} & \textbf{0.718} & \textbf{0.912} & \textbf{0.803} & 145 & 57 & 14\\
\textbf{Llama} & 0.626 & 0.817 & 0.709 & 107 & 64 & 24\\
\bottomrule
\end{tabular}
\end{table}

GPT-4o-mini demonstrates superior entity recognition coverage, identifying 77.3\% of scholarly agents present in the GT, significantly outperforming Claude (49.5\%) and Llama 3.3 (58.8\%). It identified the most entities who were expressing opinions. The perfect match rates indicate the proportion of identified entities that exactly match GT annotations. GPT-4o-mini maintains the highest accuracy at 66.0\%.

\subsection{EQ3: Evidential Reasoning Extraction Quality}\label{subsec:eq3}

\textbf{How accurately does the model capture the multi-dimensional evidential reasoning employed by scholars in their interpretations?}

Table~\ref{tab:evidence-results} presents evidence extraction performance using our custom 4-point scoring system that evaluates the accuracy of feature identification, evaluation perspective, feature classification, and polarity assessment.

\begin{table}[h]
\centering
\caption{Evidence extraction quality and coverage}
\label{tab:evidence-results}
\begin{tabular}{lcc}
\toprule
\textbf{Model} & \textbf{Mean Score (0-4)} & \textbf{Percentage Score (\%)} \\
\midrule
\textbf{Claude} & \textbf{3.87} & \textbf{96.8} \\
\textbf{GPT-4o-mini} & 3.84 & 96.0 \\
\textbf{Llama 3.3} & 3.81 & 95.3 \\
\bottomrule
\end{tabular}
\end{table}

All models demonstrate strong evidence extraction capabilities, with mean accuracies above 0.95\%. While GPT-4o-mini achieves the highest precision and recall for entities as shown in \ref{tab:entity-coverage}, Claude shows the highest evidence coverage (0.968) in Table  \ref{tab:evidence-results}.
This pattern highlights that the lower recall in identifying Cognizers by Claude returns in higher precision in downstream tasks.

\subsection{EQ4: Hypothesis and Judgment Identification}\label{subsec:eq4}

\textbf{How accurately does the model extract scholars' interpretative hypotheses and overall authenticity judgments?}

Table~\ref{tab:hypothesis-results} presents performance on extracting scholarly hypotheses about items origins and authenticity judgments.

\begin{table}[h]
\centering
\caption{Hypothesis and judgment extraction performance}
\label{tab:hypothesis-results}
\begin{tabular}{lccccc}
\toprule
\textbf{Model} & \textbf{Macro} & \textbf{Type} & \textbf{Creator} & \textbf{Date} & \textbf{Location} \\
 & \textbf{F$_1$} & \textbf{F$_1$} & \textbf{F$_1$} & \textbf{F$_1$} & \textbf{F$_1$} \\
\midrule
\textbf{Claude} & 0.655 & 0.652 & 0.638 & \textbf{0.791} & \textbf{0.923} \\
\textbf{GPT} & \textbf{0.749} & \textbf{0.845} & 0.484 & 0.595 & 0.727 \\
\textbf{Llama} & 0.694 & 0.691 & \textbf{0.712} & 0.762 & 0.727 \\
\bottomrule
\end{tabular}
\end{table}

GPT-4o-mini achieves the highest overall F$_1$-score (0.749) for hypothesis extraction, with particularly strong performance in authenticity type classification (0.845). However, the model shows weaker performance in creator hypothesis identification (0.484), suggesting challenges in extracting attribution hypotheses.

Claude demonstrates exceptional performance in geographic hypotheses (0.923 F$_1$) and temporal hypotheses (0.791 F$_1$), indicating strength in extracting location and dating alternative theories. Llama 3.3 shows the most balanced performance across hypothesis types, with particularly strong creator hypothesis extraction (0.712 F$_1$).

The variation across hypothesis types reflects the inherent complexity of scholarly reasoning, with location and date hypotheses generally more explicitly stated than creator attributions or underlying motivations.

\subsection{EQ5: Overall Discourse Representation Fidelity}\label{subsec:eq5}

\textbf{Does the complete generated KG provide an adequate representation of the scholarly debate surrounding CH Item authenticity?}

The empirical threshold, using the scores produced by G-EVAL on three well-represented articles revised manually (\textit{Posthumous Diary}, \textit{Centiloquium}, \textit{Acámbaro figures}) is set at 0.6-0.7.
This result is consistent with other evaluation findings: while the other two models demonstrate higher debate coverage overall, they are penalized for generating more FPs, resulting in lower scores.
This evaluation confirms a key pattern in our pipeline - when an entity is correctly identified as a Cognizer, their associated arguments are accurately represented. However, incorrect entity identification leads to error propagation throughout the pipeline, causing the generation of FPs in downstream components. Future iterations of the pipeline should incorporate self-consistency checks at the entity identification stage to reduce error accumulation and improve overall accuracy.
\begin{table}[h]
\centering
\caption{Per-statement Correctness (G-EVAL scores on 0-1 scale)}
\label{tab:correctness-scores}
\begin{tabular}{lccc}
\toprule
\textbf{Model} & \textbf{Mean} & \textbf{Std Dev} & \textbf{Range} \\
\midrule
\textbf{Claude} & \textbf{0.620} & 0.133 & 0.333 - 0.889 \\
\textbf{GPT} & 0.590 & 0.204 & 0.222 - 0.889 \\
\textbf{Llama} & 0.533 & 0.153 & 0.222 - 0.889 \\
\bottomrule
\end{tabular}
\end{table}

\begin{table}[h]
\centering
\caption{Overall Debate Representativeness (G-EVAL scores on 0-1 scale)}
\label{tab:representativeness-scores}
\begin{tabular}{lccc}
\toprule
\textbf{Model} & \textbf{Mean} & \textbf{Std Dev} & \textbf{Range} \\
\midrule
\textbf{Claude} & \textbf{0.607} & 0.121 & 0.333 - 0.889 \\
\textbf{GPT} & 0.580 & 0.199 & 0.222 - 0.889 \\
\textbf{Llama} & 0.523 & 0.144 & 0.222 - 0.778 \\
\bottomrule
\end{tabular}
\end{table}

\section{Discussion and Conclusions}\label{sec:discussion}
In this section, we discuss the overall performance patterns, identified bottlenecks, and potential steps to enhance the KE while answering our RQs (Section \ref{research-questions}), followed by our contributions, limitations and future steps. 

\subsection{Methodological Framework Validation}

To answer \hyperref[rq:methodological-framework]{RQ\ref*{rq:methodological-framework}}, our five-step ATR4CH methodology proves effective in coordinating LLM-based extraction with ontological frameworks. The granular evaluation demonstrates that our \textit{divide-and-conquer} methodology enables systematic refinement of individual components while maintaining system coherence. This modular evaluation strategy reveals that different models excel at different subtasks, suggesting potential for hybrid approaches that leverage each model's strengths.

The alignment between G-EVAL and other evaluations suggests that self-consistency checks throughout the pipeline (such as prompting models to evaluate their own extraction results) could reduce FPs and FNs without reducing the necessity of external validation. 

\subsection{Extraction Performance Analysis}

To answer \hyperref[rq:extraction-performance]{RQ\ref*{rq:extraction-performance}}, our evaluation reveals component-specific performance patterns across all tested models. Performance varies significantly across extraction tasks, with all models achieving high scores on metadata extraction (F$_1$-scores of 0.965-0.991), moderate performance for entity recognition (F$_1$: 0.709-0.803), strong evidence extraction capabilities (95.3-96.8\% accuracy), and more challenging hypothesis extraction (F$_1$-scores of 0.655-0.749).

This performance gradient reflects the inherent complexity of different semantic tasks rather than model-specific limitations: extracting alleged metadata proves straightforward across models, while capturing nuanced scholarly hypotheses requires more sophisticated interpretation regardless of architecture. The evidence extraction results demonstrate that contemporary LLMs can effectively capture multi-dimensional evidential reasoning, but they can do so only \textit{when they can identify the Cognizer}—this represents an error propagation problem we identified in the pipeline, as the out-of-GT outputs for evidence extraction are mostly empty or incorrect.

\subsection{Representation Fidelity and Quality Assessment}

To answer \hyperref[rq:representation-fidelity]{RQ\ref*{rq:representation-fidelity}}, the generated KGs demonstrate adequate representation of scholarly debate complexity and nuance. While the representation model proves more than adequate as already demonstrated in the BROAST catalogue \citep{amsdottoratoPasqual}, the \textit{quality} of the automatically generated KGs can still be improved.

G-EVAL scores around 0.6 indicate acceptable discourse representation quality with room for improvement. The successful capture of multi-dimensional evidential reasoning (95.3-96.8\% accuracy) shows that LLMs can handle complex semantic relationships, suggesting broader applicability to other humanities domains characterized by multi-perspectival interpretation and evidence-based reasoning. However, the model perspective on specific domain terminology and approaches requires improvement, as the G-EVAL evaluation demonstrates.

\subsection{Model Comparison and Performance Trade-offs}

To answer \hyperref[rq:model-comparison]{RQ\ref*{rq:model-comparison}}, our findings challenge the conventional assumption that larger models always perform better for complex domain tasks. The evaluation reveals distinct performance patterns across models that reflect fundamental precision-recall trade-offs rather than clear superiority based on parameter count.

Claude 3.7 Sonnet demonstrates lower recall but higher precision, being more conservative in entity classification but achieving greater accuracy in subsequent extraction steps. GPT-4o-mini shows the opposite pattern with higher recall and competitive precision, while Llama 3.3 70B falls between these approaches. Notably, as seen in Table~\ref{tab:hypothesis-results}, GPT-4o-mini performs better since it managed to correctly identify more Cognizers covered in the GT than other models, while having the least parameters of the lot.

The precision-recall trade-off has significant implications for deployment strategies. In production environments where KGs undergo human review and correction, higher recall models may be preferable since updating or deleting erroneous triples is more efficient than creating new KGs from scratch. Conversely, in real-time applications such as RAG systems where extraction occurs without human supervision, higher precision becomes critical to avoid propagating false information.

\subsection{Deployment Implications and Cost-Effectiveness}

To answer \hyperref[rq:deployment-implications]{RQ\ref*{rq:deployment-implications}}, the performance differences between models are relatively modest, while model sizes and costs differ substantially\footnote{As of May 2025, the Claude-3.7-Sonnet API has a cost of \$3/million tokens, GPT-4o-mini \$0.60/million tokens, and Llama-3.3-70B \$0.54/million tokens. The overall cost for 45 articles using the Anthropic API exceeded \$20, while for Llama-3.3.-70B and GPT-4o-mini was between \$5-10.}. This suggests that the step-by-step pipeline architecture effectively leverages the capabilities of smaller models, making deployment feasible and more cost-effective for CH institutions with varying computational budgets. 

The competitive performance of different model sizes within sequential pipelines opens two promising research directions. First, fine-tuning approaches could specifically target bottlenecks like Cognizer classification of recognized entities. Second, enhanced pre-processing using specialized tools could filter irrelevant entities before they enter the extraction pipeline. We initially considered frame recognition models for this purpose, but while these models achieve high precision in frame identification, they perform poorly in attribute classification tasks such as identifying Opinion Frame subjects, limiting their utility in entity-oriented pipelines.

The methodology's adaptability accommodates diverse institutional landscapes: smaller projects can benefit from intensive human-in-the-loop approaches with API-based models, while larger projects can leverage automated scaling through extensive annotation datasets and local deployment.

\subsection{Contributions, Limitations and Future Directions}

Our primary contributions span three interconnected domains. First, we demonstrated the practical application of the SEBI ontology using RDF-star to represent multi-perspective authenticity claims, enabling structured representation of evidence-based scholarly interpretation while preserving provenance and alternative hypotheses. Second, we introduced a comprehensive five-step methodology for building LLM-centric KE pipelines that addresses the unique challenges of humanities texts through systematic coordination of annotation models, ontological frameworks, and computational tools. The methodology's technology-agnostic design provides a replicable blueprint adaptable to varying project scales and resource constraints. Third, our technical implementation achieved practical feasibility through a sequential LLM pipeline that successfully captures scholarly reasoning including evidential features, evaluation polarities, and alternative hypotheses.

Our approach faces some limitations that can be addressed in future work. The current focus on English Wikipedia sources limits multilingual applicability, particularly important given the \textit{glocal} nature of CH scholarship. Performance on primary scholarly literature remains untested, and two key bottlenecks emerged: Cognizer classification difficulty and dependency on Wikidata linking for optimal performance.

Future work will prioritize developing multilingual extraction capabilities, implementing targeted improvements for Cognizer identification through fine-tuning or hybrid approaches, and creating user-friendly tools that enable CH practitioners to customize the extraction process with appropriate human-in-the-loop interfaces. 
Additionally, working not only with secondary literature but also with primary works from scholars would be a a relevant possible contribution. While works that try to summarize

While LLMs show promise for structuring complex scholarly debates, complete automation remains premature, suggesting that balanced human-machine collaboration represents the most viable path forward.

\section*{Acknowledgments}
For camera-ready 
Acknowledge funding sources and contributors. 

\textbf{AI Tool Disclosure:} This research employed Large Language Models (Claude Sonnet 3.7, Llama 3.3 70B, and GPT-4o-mini) as research subjects for knowledge extraction experiments, as detailed in the methodological sections of this article. Additionally, G-EVAL, an LLM-based evaluation framework, was used for assessing discourse representation quality. No AI tools were used in the writing, analysis, or interpretation of this manuscript beyond the specified cases above. The authors maintain full responsibility for the research design, methodology, data interpretation, figure creation, and all conclusions presented.

\bibliographystyle{plainnat}
\bibliography{bibliography}

@inproceedings{hartig2017foundations,
	title        = {
		Foundations of RDF\textasteriskcentered and SPARQL\textasteriskcentered:(An
		alternative approach to statement-level metadata in RDF)
	},
	author       = {Hartig, Olaf},
	year         = 2017,
	booktitle    = {
		Proceedings of the 11th Alberto Mendelzon International Workshop on
		Foundations of Data Management and the Web
	},
	location     = {Montevideo, Uruguay},
	publisher    = {CEUR-WS.org},
	series       = {CEUR Workshop Proceedings},
	volume       = 1912,
	url          = {http://ceur-ws.org/Vol-1912/paper12.pdf},
	organization = {Juan Reutter, Divesh Srivastava},
	editor       = {Juan L. Reutter and Divesh Srivastava}
}

@inproceedings{inception,
	title        = {
		The {INCE}p{TION} Platform: Machine-Assisted and Knowledge-Oriented
		Interactive Annotation
	},
	author       = {
		Klie, Jan-Christoph  and Bugert, Michael  and Boullosa, Beto  and Eckart de
		Castilho, Richard  and Gurevych, Iryna
	},
	year         = 2018,
	booktitle    = {
		Proceedings of the 27th International Conference on Computational
		Linguistics: System Demonstrations
	},
	publisher    = {Association for Computational Linguistics},
	address      = {Santa Fe, New Mexico},
	pages        = {5--9},
	url          = {https://aclanthology.org/C18-2002/},
	editor       = {Zhao, Dongyan}
}

@inproceedings{andrews2023,
	title        = {
		The Structured Assertion Record (STAR) Model for Event-based Representation
		of Historical Information
	},
	author       = {Andrews, T.},
	year         = 2023,
	booktitle    = {GrapHNR 2023},
	address      = {Mainz, Germany},
	url          = {
		https://graphentechnologien.hypotheses.org/files/2023/05/GrapHNR-2023-32-Andrews-STAR.pdf
	}
}

@inproceedings{barabucci2021,
	title        = {Supporting complexity and conjectures in cultural heritage descriptions},
	author       = {Barabucci, G. and Tomasi, F. and Vitali, F.},
	year         = 2021,
	booktitle    = {
		Proceedings of the International Conference Collect and Connect: Archives and
		Collections in a Digital Age
	},
	publisher    = {CEUR Workshop},
	pages        = {104--115}
}

@article{Tamaauskait2022DefiningAK,
	title        = {Defining a Knowledge Graph Development Process Through a Systematic Review},
	author       = {Gyt\.{e} Tamasauskait\.{e} and Paul Groth},
	year         = 2022,
	journal      = {ACM Transactions on Software Engineering and Methodology},
	volume       = 32,
	pages        = {1--40},
	url          = {https://api.semanticscholar.org/CorpusID:248435579}
}

@inproceedings{Bernasconi_Ferilli_2024,
	title        = {
		New frontiers in Digital Libraries: The trajectory of Digital Humanities
		through a computational lens
	},
	author       = {Bernasconi, E. and Ferilli, S.},
	year         = 2024,
	booktitle    = {3rd Workshop on Artificial Intelligence for Cultural Heritage (AI4CH 2024)},
	address      = {Bolzano, Italy},
	series       = {AI4CH 2024},
	doi          = {10.5281/zenodo.14923857},
	url          = {https://ai4ch.di.unito.it/},
	days         = {26--28}
}

@article{expliciting2025Giagnolini,
	title        = {
		Expliciting Contexts: Semantic Knowledge Extraction from Traditional Archival
		Descriptions
	},
	author       = {
		Giagnolini, Lucia and Schimmenti, Andrea and Bonora, Paolo and Tomasi,
		Francesca
	},
	year         = 2025,
	journal      = {Umanistica Digitale},
	volume       = 9,
	number       = 20,
	pages        = {115--144},
	doi          = {10.6092/issn.2532-8816/21229},
	url          = {https://umanisticadigitale.unibo.it/article/view/21229}
}

@article{baronicini2023,
	title        = {
		Is dc:subject enough? A landscape on iconography and iconology statements of
		knowledge graphs in the semantic web
	},
	author       = {
		Baroncini, S. and Sartini, B. and van Erp, M. and Tomasi, F. and Gangemi, A.
	},
	year         = 2023,
	journal      = {Journal of Documentation},
	volume       = 79,
	pages        = {115--136},
	doi          = {10.1108/JD-09-2022-0207}
}

@article{barone1912,
	title        = {
		Intorno alla falsificazione dei documenti ed alla critica di essi. Memoria
		letta all'Accademia Pontaniana nella tornata del 21 gennaio 1912
	},
	author       = {Barone, N.},
	year         = 1912,
	journal      = {Atti Dell'Accademia Pontaniana},
	volume       = 42,
	url          = {http://www.rmoa.unina.it/4359/}
}

@article{blau2011,
	title        = {Uncertainty and the history of ideas},
	author       = {Blau, N.},
	year         = 2011,
	journal      = {History and Theory},
	volume       = 50,
	number       = 3,
	pages        = {358--372},
	doi          = {10.1111/j.1468-2303.2011.00590.x}
}

@article{carroll2005,
	title        = {Named graphs},
	author       = {Carroll, J.J. and Bizer, C. and Hayes, P. and Stickler, P.},
	year         = 2005,
	journal      = {Journal of Web Semantics},
	volume       = 3,
	number       = 4,
	pages        = {247--267},
	doi          = {10.1016/j.websem.2005.09.001}
}

@article{daquino2020,
	title        = {
		Knowledge Representation of Digital Hermeneutics of Archival and Literary
		Sources
	},
	author       = {Daquino, Marilena and Pasqual, Valentina and Tomasi, Francesca},
	year         = 2020,
	journal      = {JLIS.It},
	volume       = 11,
	number       = 3,
	pages        = {59--76},
	doi          = {10.4403/jlis.it-12642}
}

@inproceedings{daquino2015,
	title        = {
		Historical Context Ontology (HiCO): A Conceptual Model for Describing Context
		Information of Cultural Heritage Objects
	},
	author       = {Daquino, Marilena and Tomasi, Francesca},
	year         = 2015,
	booktitle    = {Metadata and Semantics Research. MTSR 2015},
	publisher    = {Springer},
	address      = {Cham},
	series       = {Communications in Computer and Information Science},
	volume       = 544,
	doi          = {10.1007/978-3-319-24129-6_37},
	editor       = {Garoufallou, E. and Hartley, R. and Gaitanou, P.}
}

@article{dipasquale2024,
	title        = {
		On assessing weaker logical status claims in Wikidata cultural heritage
		records
	},
	author       = {
		Di Pasquale, Alessio and Pasqual, Valentina and Tomasi, Francesca and Vitali,
		Fabio
	},
	year         = 2024,
	journal      = {Semantic Web Journal}
}

@inproceedings{schick2023toolformer,
	title        = {Toolformer: Language Models Can Teach Themselves to Use Tools},
	author       = {
		Timo Schick and Jane Dwivedi-Yu and Roberto Dessi and Roberta Raileanu and
		Maria Lomeli and Eric Hambro and Luke Zettlemoyer and Nicola Cancedda and
		Thomas Scialom
	},
	year         = 2023,
	booktitle    = {Thirty-seventh Conference on Neural Information Processing Systems},
	url          = {https://openreview.net/forum?id=Yacmpz84TH}
}

@article{qin2024toollearningfoundationmodels,
	title        = {Tool Learning with Foundation Models},
	author       = {
		Qin, Yujia and Hu, Shengding and Lin, Yankai and Chen, Weize and Ding, Ning
		and Cui, Ganqu and Zeng, Zheni and Zhou, Xuanhe and Huang, Yufei and Xiao,
		Chaojun and Han, Chi and Fung, Yi Ren and Su, Yusheng and Wang, Huadong and
		Qian, Cheng and Tian, Runchu and Zhu, Kunlun and Liang, Shihao and Shen,
		Xingyu and Xu, Bokai and Zhang, Zhen and Ye, Yining and Li, Bowen and Tang,
		Ziwei and Yi, Jing and Zhu, Yuzhang and Dai, Zhenning and Yan, Lan and Cong,
		Xin and Lu, Yaxi and Zhao, Weilin and Huang, Yuxiang and Yan, Junxi and Han,
		Xu and Sun, Xian and Li, Dahai and Phang, Jason and Yang, Cheng and Wu,
		Tongshuang and Ji, Heng and Li, Guoliang and Liu, Zhiyuan and Sun, Maosong
	},
	year         = 2024,
	journal      = {ACM Comput. Surv.},
	publisher    = {Association for Computing Machinery},
	address      = {New York, NY, USA},
	volume       = 57,
	number       = 4,
	doi          = {10.1145/3704435},
	issn         = {0360-0300},
	url          = {https://doi.org/10.1145/3704435},
	issue_date   = {April 2025},
	articleno    = 101,
	numpages     = 40
}

@misc{lairgiItext2kg2024,
	title        = {
		{iText2KG}: {Incremental} {Knowledge} {Graphs} {Construction} {Using} {Large}
		{Language} {Models}
	},
	author       = {
		Lairgi, Yassir and Moncla, Ludovic and Cazabet, R\'{e}my and Benabdeslem,
		Khalid and Cl\'{e}au, Pierre
	},
	year         = 2024,
	publisher    = {arXiv},
	url          = {http://arxiv.org/abs/2409.03284},
	urldate      = {2024-09-17}
}

@inproceedings{dong2014adaptive,
	title        = {
		Adaptive Recursive Neural Network for Target-dependent Twitter Sentiment
		Classification
	},
	author       = {
		Dong, Li and Wei, Furu and Tan, Chuanqi and Tang, Duyu and Zhou, Ming and Xu,
		Ke
	},
	year         = 2014,
	booktitle    = {
		Proceedings of the 52nd Annual Meeting of the Association for Computational
		Linguistics (Volume 2: Short Papers)
	},
	publisher    = {Association for Computational Linguistics},
	address      = {Baltimore, Maryland},
	pages        = {49--54},
	url          = {https://aclanthology.org/P14-2009}
}

@book{gadamer2013,
	title        = {Truth and Method},
	author       = {Gadamer, Hans G.},
	year         = 2013,
	publisher    = {A\&C Black},
	address      = {London}
}

@article{gao2019target,
	title        = {Target-Dependent Sentiment Classification With BERT},
	author       = {Gao, Z. and Feng, A. and Song, X. and Wu, X.},
	year         = 2019,
	journal      = {IEEE Access},
	volume       = 7,
	pages        = {154290--154299},
	doi          = {10.1109/ACCESS.2019.2946594}
}

@techreport{haider2022,
	title        = {
		Verzeichnis Der Den Ober\"{o}sterreichischen Raum Betreffenden
		Gef\"{a}lschten, Manipulierten Oder Verd\"{a}chtigten Mittelalterlichen
		Urkunden
	},
	author       = {Haider, S.},
	year         = 2022,
	institution  = {Ober\"{o}sterreichisches Landesarchiv}
}

@inproceedings{hamborg2021towards,
	title        = {Towards Target-Dependent Sentiment Classification in News Articles},
	author       = {Hamborg, F. and Donnay, K. and Gipp, B.},
	year         = 2021,
	booktitle    = {Diversity, Divergence, Dialogue. iConference 2021},
	publisher    = {Springer},
	address      = {Cham},
	series       = {Lecture Notes in Computer Science},
	volume       = 12646,
	pages        = {157--169},
	doi          = {10.1007/978-3-030-71305-8_12},
	editor       = {Toeppe, K. and Yan, H. and Chu, S.K.W.}
}

@inproceedings{hartel2017,
	title        = {Il Falso Documento Del Conte Giovanni Di Moggio (875)},
	author       = {H\"{a}rtel, R.},
	year         = 2017,
	booktitle    = {
		Mue\c{c}. Societ\^{a}t Filologjiche Furlane/Societ\`{a} Filologica Friulana,
		XCIV Congr\`{e}s
	},
	address      = {Udin/Udine},
	pages        = {247--252},
	editor       = {Pugnetti, G. and Lucci, B.}
}

@techreport{lebo2013,
	title        = {PROV-O: The PROV Ontology},
	author       = {Lebo, T. and others},
	year         = 2013,
	url          = {http://www.w3.org/TR/2013/REC-prov-o-20130430/},
	institution  = {World Wide Web Consortium},
	type         = {W3C Recommendation}
}

@inproceedings{meyer2024llm,
	title        = {LLM-assisted Knowledge Graph Engineering: Experiments with ChatGPT},
	author       = {Meyer, L.-P. and Stadler, C. and others},
	year         = 2024,
	booktitle    = {
		First Working Conference on Artificial Intelligence Development for a
		Resilient and Sustainable Tomorrow. AIDRST 2023
	},
	publisher    = {Springer Vieweg},
	address      = {Wiesbaden},
	series       = {Informatik aktuell},
	pages        = {157--169},
	doi          = {10.1007/978-3-658-43705-3_8},
	editor       = {Zinke-Wehlmann, C. and Friedrich, J.}
}

@inproceedings{min2022rethinking,
	title        = {Rethinking the Role of Demonstrations: What Makes In-Context Learning Work?},
	author       = {
		Min, S. and Lyu, X. and Holtzman, A. and Artetxe, M. and Lewis, M. and
		Hajishirzi, H. and Zettlemoyer, L.
	},
	year         = 2022,
	booktitle    = {
		Proceedings of the 2022 Conference on Empirical Methods in Natural Language
		Processing
	},
	publisher    = {Association for Computational Linguistics},
	address      = {Abu Dhabi, United Arab Emirates},
	pages        = {11048--11064},
	url          = {https://aclanthology.org/2022.emnlp-main.759},
	editor       = {Goldberg, Yoav and Kozareva, Zornitsa and Zhang, Yue}
}

@inproceedings{mihindukulasooriya2023text2kgbench,
	title        = {
		Text2KGBench: A Benchmark for Ontology-Driven Knowledge Graph Generation from
		Text
	},
	author       = {Mihindukulasooriya, N. and Tiwari, S. and Fern\'{a}ndez, E.C. and Lata, K.},
	year         = 2023,
	booktitle    = {The Semantic Web – ISWC 2023},
	publisher    = {Springer},
	address      = {Cham},
	series       = {Lecture Notes in Computer Science},
	volume       = 14266,
	doi          = {10.1007/978-3-031-47243-5_14},
	editor       = {
		Payne, Terry R. and Huynh, Duy Dinh and Kim, Joongmin and Haddad, Hala and
		Afzal, Zareen and Pan, Jeff Z. and Chapman, Mark and Gandon, Fabien L. and
		Krishna, Rohit and Dumontier, Michel and Zhao, Jun
	}
}

@article{piotrowski2023uncertainty,
	title        = {Uncertainty as Unavoidable Good},
	author       = {Piotrowski, M.},
	year         = 2023,
	journal      = {Universit\"{a}t Bielefeld, Center for Uncertainty Studies (CeUS)},
	volume       = 5,
	pages        = 10,
	doi          = {10.4119/unibi/2983506},
	url          = {https://pub.uni-bielefeld.de/record/2983506}
}

@inproceedings{piotrowski2020prospects,
	title        = {Prospects for computational hermeneutics},
	author       = {Piotrowski, M. and Neuwirth, M.},
	year         = 2020,
	booktitle    = {Proceedings of the 9th AIUCD Annual Conference},
	url          = {http://amsacta.unibo.it/6316/}
}

@inproceedings{pontiki2014semeval,
	title        = {SemEval-2014 Task 4: Aspect Based Sentiment Analysis},
	author       = {
		Pontiki, M. and Galanis, D. and Pavlopoulos, J. and Papageorgiou, H. and
		Androutsopoulos, I. and Manandhar, S.
	},
	year         = 2014,
	booktitle    = {
		Proceedings of the 8th International Workshop on Semantic Evaluation (SemEval
		2014)
	},
	publisher    = {Association for Computational Linguistics},
	address      = {Dublin, Ireland},
	pages        = {27--35},
	url          = {https://aclanthology.org/S14-2004},
	editor       = {Nakov, Preslav and Zesch, Torsten}
}

@article{sartini2023,
	title        = {ICON: An Ontology for Comprehensive Artistic Interpretations},
	author       = {
		Sartini, B. and Baroncini, S. and van Erp, M. and Tomasi, F. and Gangemi, A.
	},
	year         = 2023,
	journal      = {J. Comput. Cult. Herit.},
	volume       = 16,
	number       = 3,
	pages        = {59--76},
	doi          = {10.1145/3594724}
}

@incollection{checkland_data_2006,
	title        = {Data, capta, information and knowledge},
	author       = {Peter Checkland and Sue Holwell},
	year         = 2006,
	booktitle    = {Introducing {Information} {Management}: the business approach},
	publisher    = {Elsevier},
	pages        = {47--55},
	doi          = {10.4324/9780080458397-10},
	isbn         = {0-7506-6668-4},
	language     = {English}
}

@book{valla2023,
	title        = {The treatise of Lorenzo Valla on the Donation of Constantine},
	author       = {Valla, L.},
	year         = 2023,
	publisher    = {New Haven: Yale University Press},
	url          = {https://www.gutenberg.org/ebooks/70092}
}

@misc{wang2023ischatgpt,
	title        = {Is ChatGPT a Good Sentiment Analyzer? A Preliminary Study},
	author       = {Wang, Z. and Xie, Q. and Ding, Z. and Feng, Y. and Xia, R.},
	year         = 2023,
	journal      = {ArXiv},
	volume       = {abs/2304.04339},
	url          = {https://api.semanticscholar.org/CorpusID:258048703}
}

@inproceedings{lisena_capturing_2022,
	title        = {
		Capturing the {Semantics} of {Smell}: {The} {Odeuropa} {Data} {Model} for
		{Olfactory} {Heritage} {Information}
	},
	author       = {
		Lisena, Pasquale and Schwabe, Daniel and van Erp, Marieke and Troncy,
		Rapha\"{e}l and Tullett, William and Leemans, Inger and Marx, Lizzie and
		Ehrich, Sofia Colette
	},
	year         = 2022,
	booktitle    = {The {Semantic} {Web}},
	publisher    = {Springer International Publishing},
	address      = {Cham},
	pages        = {387--405},
	isbn         = {978-3-031-06981-9},
	editor       = {
		Groth, Paul and Vidal, Maria-Esther and Suchanek, Fabian and Szekley, Pedro
		and Kapanipathi, Pavan and Pesquita, Catia and Skaf-Molli, Hala and Tamper,
		Minna
	}
}

@inproceedings{extreme_design_ontology,
	title        = {eXtreme design with content ontology design patterns},
	author       = {Presutti, Valentina and Daga, Enrico and Gangemi, Aldo and Blomqvist, Eva},
	year         = 2009,
	booktitle    = {
		Proceedings of the 2009 International Conference on Ontology Patterns -
		Volume 516
	},
	location     = {Washington DC},
	publisher    = {CEUR-WS.org},
	address      = {Aachen, DEU},
	series       = {WOP'09},
	pages        = {83–97},
	numpages     = 15
}

@article{tomasi2020,
	title        = {
		Digital humanities e organizzazione della conoscenza: una pratica di
		insegnamento nel LODLAM
	},
	author       = {Tomasi, Francesca},
	year         = 2020,
	journal      = {AIB STUDI},
	volume       = 60,
	number       = 2,
	pages        = {411--425},
	doi          = {10.2426/aibstudi-12068},
	url          = {https://aibstudi.aib.it/article/view/12068},
	language     = {italian}
}

@inproceedings{gangemi_musicbo_2024,
	title        = {
		{MusicBO}, an application of {Text2AMR2FRED} to the {Musical} {Heritage}
		domain
	},
	author       = {
		Gangemi, Aldo and Graciotti, Arianna and Marzi, Eleonora and Meloni,
		Antonello and Nuzzolese, Andrea and Presutti, Valentina and Reforgiato
		Recupero, Diego and Russo, Alessandro and Tripodi, Rocco
	},
	year         = 2024,
	booktitle    = {20th {Extended} {Semantic} {Web} {Conference}},
	publisher    = {CEUR Workshop Proceedings},
	address      = {Crete, Greece}
}

@inproceedings{petroni2019language,
	title        = {Language Models as Knowledge Bases?},
	author       = {
		Petroni, Fabio  and Rockt{\"a}schel, Tim  and Riedel, Sebastian  and Lewis,
		Patrick  and Bakhtin, Anton  and Wu, Yuxiang  and Miller, Alexander
	},
	year         = 2019,
	booktitle    = {
		Proceedings of the 2019 Conference on Empirical Methods in Natural Language
		Processing and the 9th International Joint Conference on Natural Language
		Processing (EMNLP-IJCNLP)
	},
	publisher    = {Association for Computational Linguistics},
	address      = {Hong Kong, China},
	pages        = {2463--2473},
	doi          = {10.18653/v1/D19-1250},
	url          = {https://aclanthology.org/D19-1250/},
	editor       = {Inui, Kentaro  and Jiang, Jing  and Ng, Vincent  and Wan, Xiaojun}
}

@phdthesis{amsdottoratoPasqual,
	title        = {
		The Critical Inquiry in Humanities Knowledge Graphs: Challenges, Methods,
		Innovations
	},
	author       = {Valentina Pasqual},
	year         = 2025,
	school       = {alma}
}

@misc{ontotext_graphdb_2024,
	title        = {GraphDB: Semantic Database},
	author       = {{Ontotext}},
	year         = 2024,
	url          = {https://www.ontotext.com/products/graphdb/}
}

@inproceedings{viewsariOndraszek,
	title        = {
		{eXtreme Design for Ontological Engineering in the Digital Humanities with
		Viewsari, a Knowledge Graph of Giorgio Vasari's The Lives}
	},
	author       = {
		Sarah Rebecca Ondraszek and Grischka Petri and Ulrike Blumenthal and Lisa
		Dieckmann and Etienne Posthumus and Harald Sack
	},
	year         = 2024,
	booktitle    = {
		Proceedings of the 4th Workshop on Semantic Web Technologies for Digital
		Humanities (SemDH 2024)
	},
	url          = {https://ceur-ws.org/Vol-3724/paper5.pdf}
}

@inproceedings{zaratiana2024gliner,
	title        = {
		{GL}i{NER}: Generalist Model for Named Entity Recognition using Bidirectional
		Transformer
	},
	author       = {
		Zaratiana, Urchade  and Tomeh, Nadi  and Holat, Pierre  and Charnois, Thierry
	},
	year         = 2024,
	booktitle    = {
		Proceedings of the 2024 Conference of the North American Chapter of the
		Association for Computational Linguistics: Human Language Technologies
		(Volume 1: Long Papers)
	},
	publisher    = {Association for Computational Linguistics},
	address      = {Mexico City, Mexico},
	pages        = {5364--5376},
	doi          = {10.18653/v1/2024.naacl-long.300},
	url          = {https://aclanthology.org/2024.naacl-long.300},
	editor       = {Duh, Kevin  and Gomez, Helena  and Bethard, Steven}
}

@inproceedings{abacha2025medecbenchmarkmedicalerror,
	title        = {
		{MEDEC}: A Benchmark for Medical Error Detection and Correction in Clinical
		Notes
	},
	author       = {
		Ben Abacha, Asma  and Yim, Wen-wai  and Fu, Yujuan  and Sun, Zhaoyi  and
		Yetisgen, Meliha  and Xia, Fei  and Lin, Thomas
	},
	year         = 2025,
	booktitle    = {Findings of the Association for Computational Linguistics: ACL 2025},
	publisher    = {Association for Computational Linguistics},
	address      = {Vienna, Austria},
	pages        = {22539--22550},
	doi          = {10.18653/v1/2025.findings-acl.1159},
	isbn         = {979-8-89176-256-5},
	url          = {https://aclanthology.org/2025.findings-acl.1159/},
	editor       = {
		Che, Wanxiang  and Nabende, Joyce  and Shutova, Ekaterina  and Pilehvar,
		Mohammad Taher
	}
}

@inproceedings{fewshotlearners,
	title        = {Language Models are Few-Shot Learners},
	author       = {
		Brown, Tom and Mann, Benjamin and Ryder, Nick and Subbiah, Melanie and
		Kaplan, Jared D and Dhariwal, Prafulla and Neelakantan, Arvind and Shyam,
		Pranav and Sastry, Girish and Askell, Amanda and Agarwal, Sandhini and
		Herbert-Voss, Ariel and Krueger, Gretchen and Henighan, Tom and Child, Rewon
		and Ramesh, Aditya and Ziegler, Daniel and Wu, Jeffrey and Winter, Clemens
		and Hesse, Chris and Chen, Mark and Sigler, Eric and Litwin, Mateusz and
		Gray, Scott and Chess, Benjamin and Clark, Jack and Berner, Christopher and
		McCandlish, Sam and Radford, Alec and Sutskever, Ilya and Amodei, Dario
	},
	year         = 2020,
	booktitle    = {Advances in Neural Information Processing Systems},
	publisher    = {Curran Associates, Inc.},
	volume       = 33,
	pages        = {1877--1901},
	url          = {
		https://proceedings.neurips.cc/paper\%5Ffiles/paper/2020/file/1457c0d6bfcb4967418bfb8ac142f64a-Paper.pdf
	},
	editor       = {H. Larochelle and M. Ranzato and R. Hadsell and M.F. Balcan and H. Lin}
}

@inproceedings{devlin2019bert,
	title        = {
		{BERT}: Pre-training of Deep Bidirectional Transformers for Language
		Understanding
	},
	author       = {
		Devlin, Jacob  and Chang, Ming-Wei  and Lee, Kenton  and Toutanova, Kristina
	},
	year         = 2019,
	booktitle    = {
		Proceedings of the 2019 Conference of the North {A}merican Chapter of the
		Association for Computational Linguistics: Human Language Technologies,
		Volume 1 (Long and Short Papers)
	},
	publisher    = {Association for Computational Linguistics},
	address      = {Minneapolis, Minnesota},
	pages        = {4171--4186},
	doi          = {10.18653/v1/N19-1423},
	url          = {https://aclanthology.org/N19-1423/},
	editor       = {Burstein, Jill  and Doran, Christy  and Solorio, Thamar}
}

@inproceedings{gardent2017creating,
	title        = {Creating Training Corpora for {NLG} Micro-Planners},
	author       = {
		Gardent, Claire  and Shimorina, Anastasia  and Narayan, Shashi  and
		Perez-Beltrachini, Laura
	},
	year         = 2017,
	booktitle    = {
		Proceedings of the 55th Annual Meeting of the Association for Computational
		Linguistics (Volume 1: Long Papers)
	},
	publisher    = {Association for Computational Linguistics},
	address      = {Vancouver, Canada},
	pages        = {179--188},
	doi          = {10.18653/v1/P17-1017},
	url          = {https://www.aclweb.org/anthology/P17-1017.pdf}
}

@article{sparqlAnything2024Asprino,
	title        = {
		Knowledge Graph Construction with a Fa\c{c}ade: A Unified Method to Access
		Heterogeneous Data Sources on the Web
	},
	author       = {Asprino, Luigi and Daga, Enrico and Gangemi, Aldo and Mulholland, Paul},
	year         = 2023,
	journal      = {ACM Trans. Internet Technol.},
	publisher    = {Association for Computing Machinery},
	address      = {New York, NY, USA},
	volume       = 23,
	number       = 1,
	doi          = {10.1145/3555312},
	issn         = {1533-5399},
	url          = {https://doi.org/10.1145/3555312},
	issue_date   = {February 2023},
	articleno    = 6,
	numpages     = 31
}

@inproceedings{dimou_ldow_2014,
	title        = {
		{RML:} A Generic Language for Integrated {RDF} Mappings of Heterogeneous Data
	},
	author       = {
		Dimou, Anastasia and Vander Sande, Miel and Colpaert, Pieter and Verborgh,
		Ruben and Mannens, Erik and Van de Walle, Rik
	},
	year         = 2014,
	booktitle    = {Proceedings of the 7th Workshop on Linked Data on the Web},
	series       = {CEUR Workshop Proceedings},
	volume       = 1184,
	issn         = {1613-0073},
	url          = {http://ceur-ws.org/Vol-1184/ldow2014\%5Fpaper\%5F01.pdf},
	editor       = {Bizer, Christian and Heath, Tom and Auer, S\"oren and Berners-Lee, Tim}
}

@inproceedings{rag2020Lewis,
	title        = {Retrieval-augmented generation for knowledge-intensive NLP tasks},
	author       = {
		Lewis, Patrick and Perez, Ethan and Piktus, Aleksandra and Petroni, Fabio and
		Karpukhin, Vladimir and Goyal, Naman and K\"{u}ttler, Heinrich and Lewis,
		Mike and Yih, Wen-tau and Rockt\"{a}schel, Tim and Riedel, Sebastian and
		Kiela, Douwe
	},
	year         = 2020,
	booktitle    = {
		Proceedings of the 34th International Conference on Neural Information
		Processing Systems
	},
	location     = {Vancouver, BC, Canada},
	publisher    = {Curran Associates Inc.},
	address      = {Red Hook, NY, USA},
	series       = {NIPS '20},
	isbn         = 9781713829546,
	articleno    = 793,
	numpages     = 16
}

@inproceedings{brown2020language,
	title        = {Language Models are Few-Shot Learners},
	author       = {
		Brown, Tom and Mann, Benjamin and Ryder, Nick and Subbiah, Melanie and
		Kaplan, Jared D and Dhariwal, Prafulla and Neelakantan, Arvind and Shyam,
		Pranav and Sastry, Girish and Askell, Amanda and Agarwal, Sandhini and
		Herbert-Voss, Ariel and Krueger, Gretchen and Henighan, Tom and Child, Rewon
		and Ramesh, Aditya and Ziegler, Daniel and Wu, Jeffrey and Winter, Clemens
		and Hesse, Chris and Chen, Mark and Sigler, Eric and Litwin, Mateusz and
		Gray, Scott and Chess, Benjamin and Clark, Jack and Berner, Christopher and
		McCandlish, Sam and Radford, Alec and Sutskever, Ilya and Amodei, Dario
	},
	year         = 2020,
	booktitle    = {Advances in Neural Information Processing Systems},
	publisher    = {Curran Associates, Inc.},
	volume       = 33,
	pages        = {1877--1901},
	url          = {
		https://proceedings.neurips.cc/paper\%5Ffiles/paper/2020/file/1457c0d6bfcb4967418bfb8ac142f64a-Paper.pdf
	},
	editor       = {H. Larochelle and M. Ranzato and R. Hadsell and M.F. Balcan and H. Lin}
}

@misc{dubey2024llama,
	title        = {The Llama 3 Herd of Models},
	author       = {Dubey, Abhimanyu and Jauhri, Abhinav and Pandey, Abhinav},
	year         = 2024,
	url          = {https://arxiv.org/abs/2407.21783},
	archiveprefix = {arXiv},
	description  = {The Llama 3 Herd of Models},
	eprint       = {2407.21783},
	primaryclass = {cs.AI}
}

@book{maynard2017natural,
	title        = {Natural Language Processing for the Semantic Web},
	author       = {Maynard, Diana and Bontcheva, Kalina and Augenstein, Isabelle},
	year         = 2017,
	publisher    = {Springer Cham},
	address      = {Cham, Switzerland},
	series       = {Synthesis Lectures on Data, Semantics, and Knowledge},
	pages        = {xiv, 182},
	doi          = {10.1007/978-3-031-79474-2},
	isbn         = {978-3-031-79473-5},
	url          = {https://doi.org/10.1007/978-3-031-79474-2}
}

@inproceedings{testalod2019Carriero,
	title        = {Agile Knowledge Graph Testing with TESTaLOD},
	author       = {
		Valentina Anita Carriero and Fabio Mariani and Andrea Giovanni Nuzzolese and
		Valentina Pasqual and Valentina Presutti
	},
	year         = 2019,
	booktitle    = {ISWC (Satellites)},
	pages        = {221--224},
	url          = {https://ceur-ws.org/Vol-2456/paper58.pdf},
	cdate        = 1546300800000
}

@inproceedings{he-etal-2025-evaluating-improving,
	title        = {
		Evaluating and Improving Graph to Text Generation with Large Language Models
	},
	author       = {
		He, Jie  and Yang, Yijun  and Long, Wanqiu  and Xiong, Deyi  and Gutierrez
		Basulto, Victor  and Pan, Jeff Z.
	},
	year         = 2025,
	booktitle    = {
		Proceedings of the 2025 Conference of the Nations of the Americas Chapter of
		the Association for Computational Linguistics: Human Language Technologies
		(Volume 1: Long Papers)
	},
	publisher    = {Association for Computational Linguistics},
	address      = {Albuquerque, New Mexico},
	pages        = {10219--10244},
	doi          = {10.18653/v1/2025.naacl-long.513},
	isbn         = {979-8-89176-189-6},
	url          = {https://aclanthology.org/2025.naacl-long.513/},
	editor       = {Chiruzzo, Luis  and Ritter, Alan  and Wang, Lu}
}

@inproceedings{papineni2002bleu,
	title        = {{BLEU}: a method for automatic evaluation of machine translation},
	author       = {Papineni, Kishore and Roukos, Salim and Ward, Todd and Zhu, Wei-Jing},
	year         = 2002,
	booktitle    = {
		Proceedings of the 40th Annual Meeting of the Association for Computational
		Linguistics
	},
	publisher    = {Association for Computational Linguistics},
	address      = {Philadelphia, Pennsylvania, USA},
	pages        = {311--318}
}

@inproceedings{banerjee2005meteor,
	title        = {
		{METEOR}: An automatic metric for {MT} evaluation with improved correlation
		with human judgments
	},
	author       = {Banerjee, Satanjeev and Lavie, Alon},
	year         = 2005,
	booktitle    = {
		Proceedings of the ACL Workshop on Intrinsic and Extrinsic Evaluation
		Measures for Machine Translation and/or Summarization
	},
	publisher    = {Association for Computational Linguistics},
	address      = {Ann Arbor, Michigan},
	pages        = {65--72}
}

@article{yuan2021bartscore,
	title        = {{BartScore}: Evaluating generated text as text generation},
	author       = {Yuan, Weizhe and Neubig, Graham and Liu, Pengfei},
	year         = 2021,
	journal      = {Advances in Neural Information Processing Systems},
	volume       = 34,
	pages        = {27263--27277}
}

@inproceedings{popovic2015chrf,
	title        = {{chrF}: character n-gram {F}-score for automatic {MT} evaluation},
	author       = {Popovi{\'c}, Maja},
	year         = 2015,
	booktitle    = {Proceedings of the Tenth Workshop on Statistical Machine Translation},
	publisher    = {Association for Computational Linguistics},
	address      = {Lisbon, Portugal},
	pages        = {392--395}
}

@inproceedings{geval-evaluation,
	title        = {{G}-Eval: {NLG} Evaluation using Gpt-4 with Better Human Alignment},
	author       = {
		Liu, Yang  and Iter, Dan  and Xu, Yichong  and Wang, Shuohang  and Xu,
		Ruochen  and Zhu, Chenguang
	},
	year         = 2023,
	booktitle    = {
		Proceedings of the 2023 Conference on Empirical Methods in Natural Language
		Processing
	},
	publisher    = {Association for Computational Linguistics},
	address      = {Singapore},
	pages        = {2511--2522},
	doi          = {10.18653/v1/2023.emnlp-main.153},
	url          = {https://aclanthology.org/2023.emnlp-main.153/},
	editor       = {Bouamor, Houda  and Pino, Juan  and Bali, Kalika}
}

@article{info12120503,
	title        = {
		CIDOC2VEC: Extracting Information from Atomized CIDOC-CRM Humanities
		Knowledge Graphs
	},
	author       = {El-Hajj, Hassan and Valleriani, Matteo},
	year         = 2021,
	journal      = {Information},
	volume       = 12,
	number       = 12,
	doi          = {10.3390/info12120503},
	issn         = {2078-2489},
	url          = {https://www.mdpi.com/2078-2489/12/12/503},
	article-number = 503
}

@inproceedings{morales_tirado_musical_2023,
	title        = {
		Musical {Meetups}: a {Knowledge} {Graph} approach for {Historical} {Social}
		{Network} {Analysis}
	},
	author       = {
		Morales Tirado, Alba and Carvalho, Jason and Mulholland, Paul and Daga,
		Enrico
	},
	year         = 2023,
	booktitle    = {
		{ESWC} 2023 {Workshops} and {Tutorials}. {Semantic} {Methods} for {Events}
		and {Stories} ({SEMMES})
	},
	publisher    = {CEUR Workshop Proceedings (CEUR-WS.org)},
	volume       = 3443,
	url          = {https://ceur-ws.org/Vol-3443/ESWC\%5f2023\%5fSEMMES\%5fMeetups-CR.pdf},
	editor       = {
		Alam, Mehwish and Trojahn, Cassia and Hertling, Sven and Pesquita, Catia and
		Aebeloe, Christian and Aras, Hidir and Azzam, Amr and Cano, Juan and {John
		Domingue} and Gottschalk, Simon and Hartig, Olaf and Hose, Katja and {Sabrina
		Kirrane} and Lisena, Pasquale and Osborne, Francesco and Rohde, Philipp and
		{Luc Steels} and Taelman, Ruben and Third, Aisling and Tiddi, Ilaria and
		{Rima T\"{u}rker}
	}
}

@inproceedings{Meloni2017AMR2FREDAT,
	title        = {
		AMR2FRED, A Tool for Translating Abstract Meaning Representation to
		Motif-Based Linguistic Knowledge Graphs
	},
	author       = {Antonello Meloni and Diego Reforgiato Recupero and Aldo Gangemi},
	year         = 2017,
	booktitle    = {Extended Semantic Web Conference},
	url          = {https://api.semanticscholar.org/CorpusID:34725770}
}

@article{KE_using_LLMs,
	title        = {{Knowledge Engineering Using Large Language Models}},
	author       = {Allen, Bradley P. and Stork, Lise and Groth, Paul},
	year         = 2023,
	journal      = {Transactions on Graph Data and Knowledge},
	publisher    = {Schloss Dagstuhl -- Leibniz-Zentrum f{\"u}r Informatik},
	address      = {Dagstuhl, Germany},
	volume       = 1,
	number       = 1,
	pages        = {3:1--3:19},
	doi          = {10.4230/TGDK.1.1.3},
	issn         = {2942-7517},
	url          = {https://drops.dagstuhl.de/entities/document/10.4230/TGDK.1.1.3},
	urn          = {urn:nbn:de:0030-drops-194777}
}

@misc{trends2024KGs,
	title        = {
		Research {Trends} for the {Interplay} between {Large} {Language} {Models} and
		{Knowledge} {Graphs}
	},
	author       = {
		Khorashadizadeh, Hanieh and Amara, Fatima Zahra and Ezzabady, Morteza and
		Ieng, Fr\'{e}d\'{e}ric and Tiwari, Sanju and Mihindukulasooriya, Nandana and
		Groppe, Jinghua and Sahri, Soror and Benamara, Farah and Groppe, Sven
	},
	year         = 2024,
	publisher    = {arXiv},
	doi          = {10.48550/arXiv.2406.08223},
	url          = {http://arxiv.org/abs/2406.08223},
	urldate      = {2024-09-20}
}

@article{ringwald_learning_2024,
	title        = {
		Learning {Pattern}-{Based} {Extractors} from {Natural} {Language} and
		{Knowledge} {Graphs}: {Applying} {Large} {Language} {Models} to {Wikipedia}
		and {Linked} {Open} {Data}
	},
	author       = {Ringwald, C\'{e}lian},
	year         = 2024,
	journal      = {Proceedings of the AAAI Conference on Artificial Intelligence},
	volume       = 38,
	number       = 21,
	pages        = {23411--23412},
	doi          = {10.1609/aaai.v38i21.30406},
	issn         = {2374-3468},
	url          = {https://ojs.aaai.org/index.php/AAAI/article/view/30406},
	urldate      = {2024-09-20}
}

@article{kumar_k-lm_2022,
	title        = {
		K-{LM}: {Knowledge} {Augmenting} in {Language} {Models} {Within} the
		{Scholarly} {Domain}
	},
	author       = {
		Kumar, Vivek and Reforgiato Recupero, Diego and Helaoui, Rim and Riboni,
		Daniele
	},
	year         = 2022,
	journal      = {IEEE Access},
	volume       = 10,
	pages        = {91802--91815},
	doi          = {10.1109/ACCESS.2022.3201542},
	issn         = {2169-3536}
}

@article{Doerr_2003,
	title        = {
		The CIDOC Conceptual Reference Module: An Ontological Approach to Semantic
		Interoperability of Metadata
	},
	author       = {Doerr, Martin},
	year         = 2003,
	journal      = {AI Magazine},
	volume       = 24,
	number       = 3,
	pages        = 75,
	doi          = {10.1609/aimag.v24i3.1720},
	url          = {https://ojs.aaai.org/aimagazine/index.php/aimagazine/article/view/1720},
	abstractnote = {
		This article presents the methodology that has been successfully used over
		the past seven years by an interdisciplinary team to create the International
		Committee for Documentation of the International Council of Museums (CIDOC)
		CONCEPTUAL REFERENCE MODEL (CRM), a high-level ontology to enable information
		integration for cultural heritage data and their correlation with library and
		archive information. The CIDOC CRM is now in the process to become an
		International Organization for Standardization (ISO) standard. This article
		justifies in detail the methodology and design by functional requirements and
		gives examples of its contents. The CIDOC CRM analyzes the common
		conceptualizations behind data and metadata structures to support data
		transformation, mediation, and merging. It is argued that such ontologies are
		propertycentric, in contrast to terminological systems, and should be built
		with different methodologies. It is demonstrated that ontological and
		epistemological arguments are equally important for an effective design, in
		particular when dealing with knowledge from the past in any domain. It is
		assumed that the presented methodology and the upper level of the ontology
		are applicable in a far wider domain.
	}
}

@techreport{ifla2017frbr,
	title        = {
		Definition of FRBRoo: A Conceptual Model for Bibliographic Information in
		Object-Oriented Formalism
	},
	author       = {{IFLA Working Group on FRBR/CRM Dialogue}},
	year         = 2017,
	url          = {https://repository.ifla.org/handle/20.500.14598/659},
	institution  = {International Federation of Library Associations and Institutions (IFLA)},
	version      = {2.4}
}


\end{document}